\documentclass[10pt,twocolumn,letterpaper]{article}

\usepackage{cvpr}      % To produce the REVIEW version
\usepackage[accsupp]{axessibility}
\usepackage{graphicx}
\usepackage{amsmath}
\usepackage{amssymb}
\usepackage{booktabs}
\usepackage{multirow}
\usepackage{color}

\usepackage[pagebackref,breaklinks,colorlinks]{hyperref}

\usepackage[capitalize]{cleveref}
\crefname{section}{Sec.}{Secs.}
\Crefname{section}{Section}{Sections}
\Crefname{table}{Table}{Tables}
\crefname{table}{Tab.}{Tabs.}

\def\cvprPaperID{1403} % *** Enter the CVPR Paper ID here
\def\confName{CVPR}
\def\confYear{2022}

\begin{document}
	
	%%%%%%%%% TITLE - PLEASE UPDATE
	\title{Camera-Conditioned Stable Feature Generation for Isolated Camera Supervised Person Re-IDentification}
	% \title{Stable Camera-Conditioned Feature Generation for Isolated Camera Supervised Person Re-IDentification}
	%\title{Stable Camera-Conditioned Feature Generation for Isolated Camera Person Re-IDentification}
	%\title{Stable Camera-Conditioned Feature Generation for Single Camera Supervised Person Re-IDentification}
	%\title{Discriminative Stable Cross-Camera Feature Generation for Distant Scenes Intra-Camera Supervised Person Re-IDentification}
	
% 	\author{ Chao Wu$^{1}$, Xiaobin Chang$^{1,2,3}$\thanks{indicates corresponding author.}\ \ , Wenhang Ge$^{4}$, Ancong Wu$^{4}$
%     \\
    \author{ Chao Wu$^{1}$, Wenhang Ge$^{4}$, Ancong Wu$^{4}$, Xiaobin Chang$^{1,2,3}$\thanks{indicates corresponding author.}\ \ 
    \\
    {\small $^1$School of Artificial Intelligence, Sun Yat-sen University, China}\\ {\small $^2$Guangdong Key Laboratory of Big Data Analysis and Processing, Guangzhou 510006, P.R.China} \\ 
    {\small $^3$Key Laboratory of Machine Intelligence and Advanced Computing, Ministry of Education, China} \\ 
    {\small $^4$School of Computer Science and Engineering, Sun Yat-sen University, China} 
    \\
    % {\small wuch76@mail2.sysu.edu.cn,\ changxb3@mail.sysu.edu.cn,\ gewh@mail2.sysu.edu.cn,\ wuanc@mail.sysu.edu.cn}}
    {\small wuch76@mail2.sysu.edu.cn,\ gewh@mail2.sysu.edu.cn,\ wuanc@mail.sysu.edu.cn,\ changxb3@mail.sysu.edu.cn}}

	\maketitle
	
	%%%%%%%%% ABSTRACT
	\begin{abstract}
		To learn camera-view invariant features for person Re-IDentification (Re-ID), the cross-camera image pairs of each person play an important role.
		However, such cross-view training samples could be unavailable under the ISolated Camera Supervised (ISCS) setting, e.g., a surveillance system deployed across distant scenes.
		To handle this challenging problem, a new pipeline is introduced by synthesizing the cross-camera samples in the feature space for model training. 
		Specifically, the feature encoder and generator are end-to-end optimized under a novel method, Camera-Conditioned Stable Feature Generation (CCSFG). Its joint learning procedure raises concern on the stability of generative model training. Therefore, a new feature generator, $\sigma$-Regularized Conditional Variational Autoencoder ($\sigma$-Reg.~CVAE), is proposed with theoretical and experimental analysis on its robustness.
		Extensive experiments on two ISCS person Re-ID datasets demonstrate the superiority of our CCSFG to the competitors. \footnote{\url{https://github.com/ftd-Wuchao/CCSFG}}

	\end{abstract}
	
	%%%%%%%%% BODY TEXT
	\section{Introduction}
	
\begin{figure}[t]
		\centering
		\includegraphics[width=0.39\textwidth]{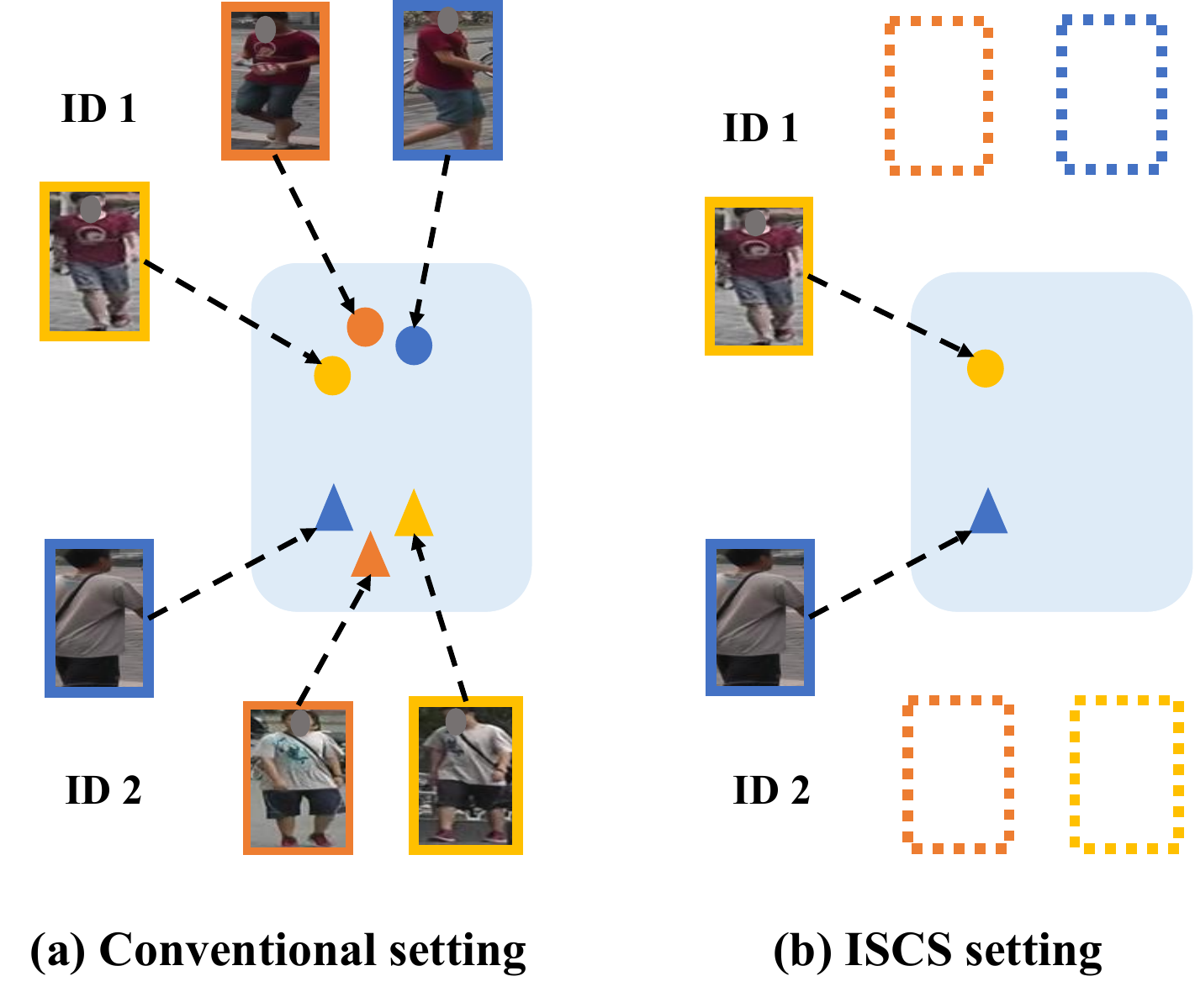}
		\caption{
			Illustrations of the training samples under different person Re-ID settings. The light-blue areas indicate the feature space. Different shapes corresponding to identities. Different colors mean under different cameras.
			(a) Cross-camera person images are available under the conventional settings.
			(b) No cross-camera image pairs under the ISCS setting for model training.
			% conventional setting, ISCS setting and our motivation. Different colors indicate person images and features from different cameras. Solid shape denotes real features, and hollow shape denotes generated features. (a) In the conventional setting, the most existing methods are heavily rely on the cross-camera paired data to learn discriminative feature. (b) In the ISCS setting, there are no cross-camera paired data, which causes a significant challenge for learning camera-unrelated feature. (c) Our motivation is to generate cross-camera feature in real time during training, to solve the problem of cross-camera paired data missing.
		}
		\label{fig:introduction}
	\end{figure}

	Person re-identification (Re-ID) aims to retrieve the same person across different cameras in a surveillance network. Extracting the discriminative view-invariant features of person images play a central role for the Re-ID task.
	With the cross-camera images of each person available during training, existing methods have made great progress under different settings, e.g., the supervised~\cite{PCB, AGW, chang2018multi, wang2018person} and the unsupervised~\cite{li2020joint, dai2021dual, wang2021central, li2021a}. The importance of cross-camera samples for model training is also demonstrated.
	However, such cross-camera person images are not guaranteed during training under some realistic scenarios.
	For example, a surveillance system is needed to re-identify a person across distant scenes, e.g., different cities, and each camera is isolated. It is too expensive to collect sufficient cross-camera person images for model training. A more applicable solution is exploiting the large amount of camera-specific images of different persons instead.
	As the cross-camera image pairs no longer exist during training, many existing methods~\cite{bagtrick,AGW,PCB,HHL} fail to obtain the ideal performance on such data.
	This challenging person Re-ID setting, called ISolated Camera Supervised (ISCS), is first proposed by~\cite{SCT} as Single-Camera-Training (SCT).
	The comparison between different settings is shown in \cref{fig:introduction}.

	% MCNL\cite{SCT} and CCFP\cite{ccfp}
	
\begin{figure}[t]
		\centering
		\includegraphics[width=0.39\textwidth]{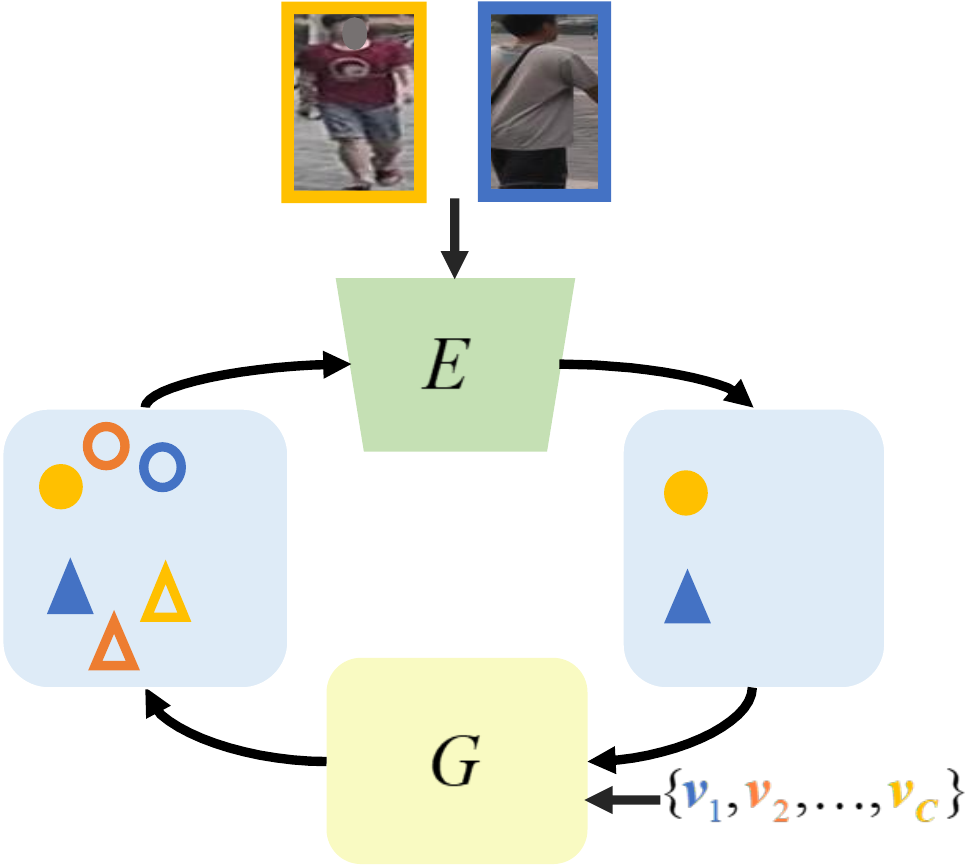}
		\caption{Our pipeline for the ISCS setting. The dots with different shapes represent the features of different IDs. Different colors mean different camera views $\{v_1,v_2,\dots,v_C\}$.
			% {\color{red}Different shapes and colors denote the different identities and features from different cameras, respectively. $\{v_1,v_2,\dots,v_C\}$ represents different camera label of the input generator G. } 
			The feature samples conditioned on different camera views are generated by the generator G, shown as the non-solid dots.
			% across different cameras are generated by generator G.
			The cross-camera pairs are thus available for better training of the encoder E.}
		\label{fig:pipeline}
	\end{figure}

	To handle the challenging ISCS settings, existing methods~\cite{SCT, ccfp} explicitly align the feature distributions across cameras with new losses and network architectures.
	In this paper, we follow an alternative pipeline based on generation. The motivation behind this is rather straightforward: As the cross-camera samples play an important role in person Re-ID model training while such paired images do not exist under the ISCS setting, the missing camera view data can be compensated by the generated ones.
	Specifically, the cross-camera samples are generated in the feature space rather than as images with two considerations. 
	Firstly, it takes great efforts of the generative model to capture details, e.g, backgrounds and illuminations, to improve the visual quality of images. The payoffs of such efforts may not directly reflect on Re-ID and the not ideal generated images can even harm the performance.
	% generating person images across different cameras is challenging.
	On the contrary, the feature generation is not distracted by visual quality and is more concentrated on introducing camera-view information while preserving the discriminative power of the generated samples.
	% The cross-camera samples can be generated in either the image space or the feature space. And the feature space is chosen in this work with the following reasons.
	With the camera-conditioned
	% {\color{red}camera-conditioned (we expect to generate cross-camera features corresponding to existing features, called camera-conditioned feature generation ) } 
	features of different persons generated, the cross-camera samples are recovered and can be used to train a better encoder of person images.
	To sum up, a new pipeline is introduced to handle the ISCS setting by synthesizing the cross-camera samples for better encoder training, as illustrated in \cref{fig:pipeline}.
	
	To instantiate the pipeline above,
	a novel method, Camera-Conditioned Stable Feature Generation (CCSFG), is proposed.
	A common CNN backbone is used as the image encoder $E$ and a camera conditioned variational autoencoder (CVAE) as the feature generator $G$.
	As the encoder $E$ and the generator $G$ are not ideal at first, they should be jointly optimized for iterative improvements.
	% iterative improvements between them can be achieved via end-to-end joint learning.
	% Such $E$ and $G$ should be jointly optimized due to their strong dependencies.
	% There are strong dependencies between the image encoder $E$ and feature generator $G$ during training.
	On the one hand, with the more reliable features conditioned on cameras generated by $G$, the person appearance features from $E$ can be more discriminative and less variant across cameras.
	% with the more reliable cross-camera features generated by G, the person appearance features from $E$ are more discriminative and less variant to different views.
	On the other hand, with the more discriminative features from $E$, the generator $G$ can be more focused on capturing the camera information.
	% As the encoder $E$ and the generator $G$ are not ideal at first, iterative improvements between them can be achieved via end-to-end joint learning.
	% they can be jointly learned to synchronize the optimizations for iterative improvements.
	% joint learning is used to synchronize the optimizations of $E$ and $G$ for iterative improvements.
	However, this joint learning procedure forms an obstacle in training the generator $G$.
	% However, this learning procedure forms an obstacle in generative model training.
	% this learning procedure introduces a challenge for generative model training.
	% such a procedure is fragile and easily collapsed due to the unstable generative model training.
	The input of $G$ is the output of encoder $E$ that is still under training. Therefore, the enlarging dynamic variance of such input causes instability in training $G$ and eventually leads to collapsed learning of the whole model.
	% Different from the conventional generative models sampling inputs from static distributions or datasets, the inputs for generator $G$ in CCSFG are the outputs of encoder $E$ that is still under training. Therefore, the huge dynamic variance of such inputs cause instability in training $G$ and eventually lead to collapsed learning.
	To handle this issue, a novel generative model, $\sigma$-Regularized CVAE ($\sigma$-Reg.~CVAE), is proposed with a simple yet effective solution based on feature normalization and used as the generator $G$.
	% To handle this issue, a novel generative model, $\sigma$-Regularized CVAE ($\sigma$-Reg.~CVAE), is proposed. The solution provided by $\sigma$-Reg.~CVAE is based on normalization, which is simple yet effective.
	More importantly, we provide the theoretical analysis and demonstrate it with experiments.
	% its theoretical analysis is provided with experiments for demonstration.
	
	% In this paper, we propose a novel feature generation pipeline, named Camera-Conditioned Stable Feature Generation (CCSFG) for generating cross-camera features, which joints feature learning and feature generation in an single network. Specifically, given no cross-camera paired data, we can get the pedestrian feature by the encoder (Resent-50). Due to the cross-camera paired data missing, it is difficult to learn camera-unrelated features. Therefore, we proposed $\sigma$ Regularized  Conditional Variational Autoencoders ($\sigma$-Reg.~CVAE) to generate distinguishing features for identity and camera view. By pulling the distance between real features and the corresponding cross-camera generated features, it makes the learned features is not related to camera view. Moreover, since the input of $\sigma$-Reg.~CVAE is features, which is not fixed in the training process, the conventional CVAE model occurs posterior collapse easily. To solve this problem, we propose an easily and effective method, called Input Feature Normalization (IFN), to keep the model stable training. Extensive experiments demonstrate the performance superiority of our model over the state-of-the-art alternative methods in the ISCS setting on Market-SCT, Duke-SCT, and MSMT-SCT datasets. 
	
	The main contributions of this paper are in three-fold.
	(1) To handle the challenging ISCS person Re-ID problem, a novel pipeline is proposed to explicitly generate the cross-view samples in the feature space for better encoder learning.
	(2) Following the pipeline above, a novel method, CCSFG, is instantiated. The encoder $E$ and generator $G$ are jointly optimized for iterative improvements.
	(3) To achieve stable joint learning in CCSFG, a novel generative model, $\sigma$-Reg.~CVAE, is proposed with detailed analysis provided.
	% for the stable joint learning in CCSFG. Detailed 
	The effectiveness of the proposed CCSFG is demonstrated by its state-of-the-art performance on two ISCS person Re-ID benchmark datasets.
	
	% The main contributions of this paper can be summarized in three aspects:
	% \begin{itemize}
	%     \item We provide the pipeline that can generate pedestrian features under the specified camera view to solve the problem of cross-camera paired data missing effectively. 
	
	%     \item We propose a novel unified method for ISCS person Re-ID, which is able to joint feature learning and generation in an end-to-end trainable network..
	%     \item We introduce input feature normalization technology to maintain the stability of generative model training with theoretical analysis.
	%     %\item We achieve a new state of the art on three benchmarks datasets and demonstrate the effectiveness and efficiency of our approach through ablation studies.
	% \end{itemize}

	\section{Related Work}
	
	\noindent\textbf{Person Re-ID Settings.}\quad
	To study the varied application scenarios of person re-id, different benchmark settings have been proposed for research.
	The person images in a dataset are usually assumed to be captured from a surveillance network with adjacent cameras but disjoint monitoring areas.
	Under the \textbf{supervised setting}~\cite{xiao2016learning, chang2018multi, PCB, AGW, MGN},  their identities are elaborated labeled and aligned across different cameras as supervision.
	The \textbf{unsupervised setting}~\cite{zhao2013unsupervised, fan2018unsupervised, li2020joint, dai2021dual,wang2021central} is more challenging than the supervised one by abandoning all the ID labels for model training.
	To help learning a model on the unsupervised target dataset, the extra source labeled data is available under the \textbf{unsupervised domain adaptation (UDA)} setting~\cite{he2018deep, zhai2020ad, song2020unsupervised, yang2021joint, isobe2021towards}.
	Moreover, the \textbf{intra-camera supervised (ICS)} setting~\cite{TAULD, UGA, MTML, TSSL, PCSL} provides the camera-specific ID labels and without a global correspondence across cameras.
	All these settings are with cross-camera images of each person for model training. Their differences lie in the supervision extent and manners.
	The recent proposed \textbf{ISolated Camera Supervised (ISCS)} person Re-ID setting~\cite{SCT, ccfp} focuses on a distinctive scenario where no cross-camera person images are available for model training.
	Therefore, to learn the view-invariant models, existing methods~\cite{SCT, ccfp} handle this challenging setting with the alignment losses on the data distributions rather than the sample pairs of different cameras.
	In this paper, the alternative pipeline based on generation is proposed under a simple and sound motivation: to recover the crucial cross-camera samples and use them for enhancing the model training.
	To synthesis the person images under new camera views, existing generative methods, e.g., HHL~\cite{HHL}, can be exploited.
	Our proposed method, CCSFG, is based on cross-camera feature generation instead. As a unified model, its image encoder $E$ and feature generator $G$ are end-to-end optimized for mutual improvement. To achieve stable joint learning, a novel feature generator, $\sigma$-Reg.~CVAE, is proposed.

	\noindent\textbf{Generative Models.}\quad
	The Variational Autoencoders (VAE)~\cite{VAE} and Generative Adversarial Networks (GAN)~\cite{GAN} are two widely exploited generative methods for computer vision problems, such as medical image segmentation~\cite{shin2018medical, baumgartner2019phiseg}, latent representations disentanglement~\cite{donahue2018semantically, gulrajani2016pixelvae} and image background modeling\cite{born2015background, li2019supervae}.
	The GAN and VAE based methods also play important role in person Re-ID problems.
	Different GAN-based methods have been proposed to augment the training person images under the supervised setting~\cite{Liu2018PoseTP, Qian2018PoseNormalizedIG, Zhong2018CameraSA, zheng2019joint}.
	To bridge the domain gaps of different datasets under the UDA setting, the GAN-based methods~\cite{deng2018image, liu2019adaptive, ding2021deeply, huang2019sbsgan, HHL} are proposed to transfer person image styles across domains.
	Existing VAE-based methods~\cite{Pu2020DualGV, Ren2021HAVANAHA} for person Re-ID mainly focus on disentangled representation learning rather than explicitly generating samples.
	To our best knowledge, the proposed $\sigma$-Reg.~CVAE is the first VAE-based feature generator for the ISCS person Re-ID.
	% The missing cross-camera samples under the ISCS setting can be recovered with the camera-view conditioned synthesized features by $\sigma$-Reg.~CVAE.
	The generator $\sigma$-Reg.~CVAE and the encoder are unified under our CCSFG method for joint learning.
	% Different from the conventional generative models~\cite{VAE,GAN} sampling inputs from static distributions or datasets
	However, the input for generator in CCSFG is the output of encoder that is still under training. Therefore, the huge dynamic variance of such inputs causes instability in training $G$ and eventually leads to collapsed learning~\cite{CollapseICML2020, Rybkin2021SimpleAE, takida2021preventing}.
	With theoretical and experimental analysis, a simple yet effective solution to this issue is proposed and incorporated by $\sigma$-Reg.~CVAE.

	\section{Methodology}
	
	\subsection{Isolated Camera Supervised Person Re-ID}
	%{\color{red} We use mathematical formulas to describe this setting, highlighting the problem of no cross-camera paired data.}
	
	The training set is denoted as $\mathcal{D} = \{(\boldsymbol{x}_n,y_n,c_n)\}_{n=1}^{|\mathcal{D}|}$, where each training sample is a triplet with the person image $\boldsymbol{x}_n$, its identity label $y_n\in\{p_1,\dots,p_M\}$ and the camera label $c_n \in \{v_1,\dots,v_C\}$. $M$ and $C$ denote the total numbers of different identities and camera views for training respectively.
	Under the ISCS setting, the cross-camera images of the same person do not exist in the training set, i.e., $\forall i,j \in \{1, ..., |\mathcal{D}|\}$, if $c_i \neq c_j$, then $y_i \neq y_j$.
	The testing protocol follows the regular routine. Given a query image of a pedestrian, a Re-ID model aims to retrieve the images of the same person from the gallery set.
	
	% We denote the training data as $\mathcal{D} = \{(x_n,y_n,c_n)\}_{n=1}^{|\mathcal{D}|}$, where $x_n$ is the n-th person image in training set. Each person image $x_n$ is annotated with an identity label $y_n\in\{p_1,\dots,p_I\}$ and a camera label $c_n \in \{v_1,\dots,v_C\}$. $I$ and $C$ denote the number of identities and cameras in training set. In ISCS setting, the cross-camera image pairs are no longer exist in training set, i.e., there does not exist $x_m$, $x_n$ and $c_n \neq c_m$ that satisfy $y_m = y_n$.
	% Given a query of a certain pedestrian, the person Re-ID aims to match the same person by finding a ranked list of images from the gallery set according to the similarity. The existence of cross-camera paired data missing prevents the training of conventional supervised Re-ID models, thus a new effective Re-ID method is needed. 
	
	% As show in \cref{fig:framework}, we proposed the Camera-Conditioned Stable Feature Generation method (CCSFG), which is composed of Encoder $E$ and Generator $G$. $E$ is a  ReID model to produce feature vectors. $G$ is a generative model $\sigma$-Reg.~CVAE to generate the cross-camera samples in the feature space for model training. $E$ and $G$ are end-to-end trained under a unified framework. The training objective of CCSFG could be formulated as:
	% \begin{equation}
	%     \mathop{\min}\limits_{G,E}{L}_{(G,E)}(x,y,c) = \mathcal{L}_{G|E}(x,y,c) + \mathcal{L}_{E|G}(x,y),
	% \end{equation}
	% where $\boldsymbol{x}$ denotes an input image, $y$ and $c$ denote the identity label and camera label, respectively.
	
	\begin{figure}[t]
		\centering
		\includegraphics[width=0.45\textwidth]{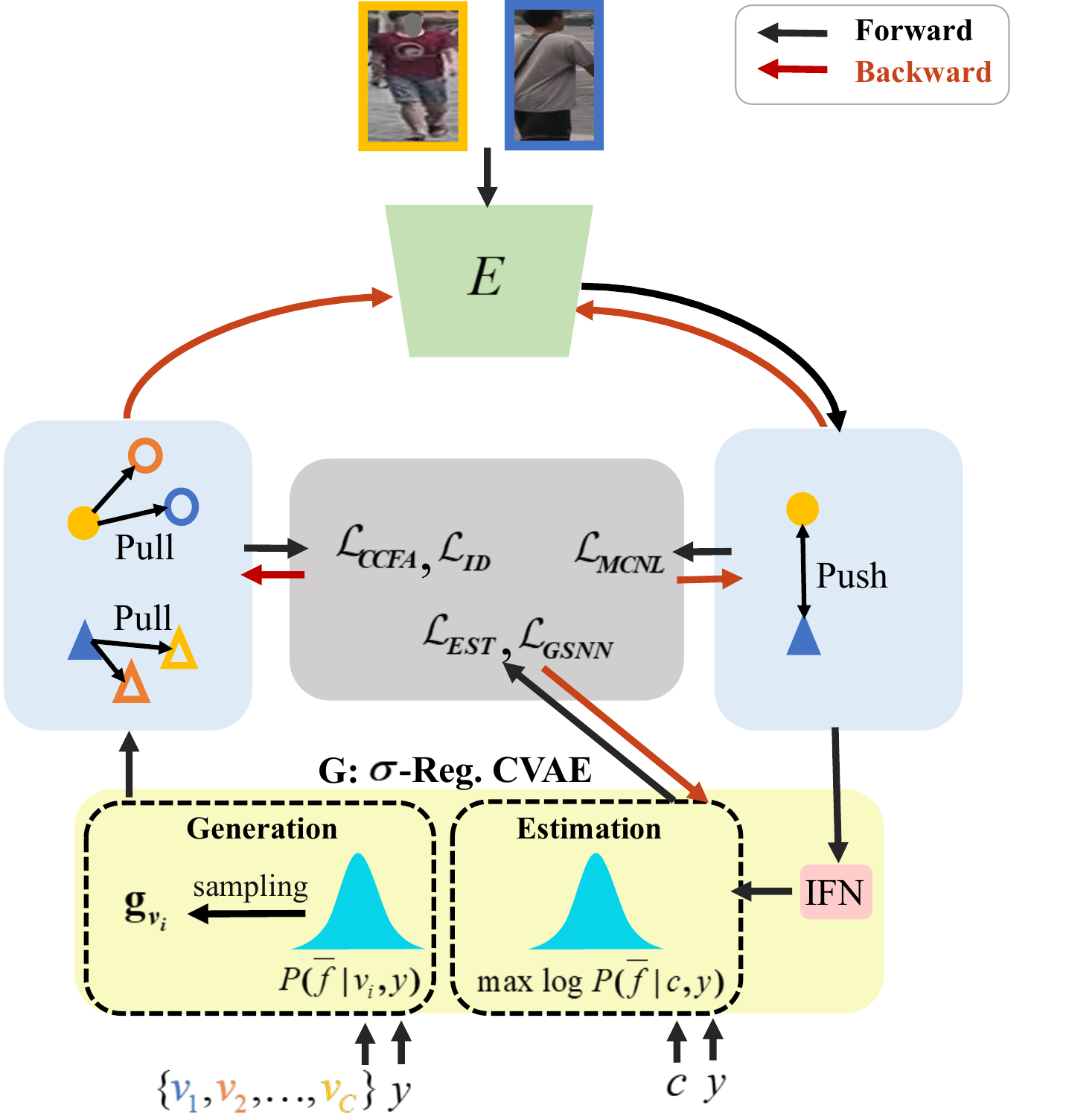}
		\caption{Overview of the Camera-Conditioned Stable Feature Generation (CCSFG) method. $E$ denotes the encoder to extract features from pedestrian images.
			The features $\{\boldsymbol{g}_{v_i}\}_{i=1}^{C}$ are generated by G conditioning on the identity $y$ and camera views $\{v_i\}_{i=1}^{C}$.
			IFN denotes the normalization module of generator G.
			% {\color{red} $E$ denotes the encoder to extract the feature $f$ from the pedestrian image. IFN denotes the input feature normalization to keep the stability of model training, which will convert $f$ to $\bar{f}$. The generator $G$ uses the identity label $y$ and camera view $\{v_i\}_{i=1}^{C}$ as input to generate camera-conditioned features $\{\boldsymbol{g}_{v_i}\}_{i=1}^{C}$.}
			%{\color{red}which consists of the encoder $E$ and the generator $G$. Fristly, we extract the person feature $\boldsymbol{f}$ by $E$ and apply MCNL loss ($\mathcal{L}_{MCNL}$) for the feature similarity learning. For the stable joint learning of CCSFG , $f$ is converted to $\bar{f}$ by the input feature normalization (IFN) function. Afterwards, the feature $\bar{f}$, identity label $y$ and camera label $c$ is used as the input of $G$ to the to maximize the conditional log-likelihood $\log P(\boldsymbol{\bar{f}}|c,y)$ by the
			%$\mathcal{L}_{EST}$ and $\mathcal{L}_{GSNN}$. Meanwhile, $G$ generate the camera-conditioned features $\{g_{v_i}\}_{i=1}^{C}$ to help $E$ learn camrea-unrelated features. Finally, the cross entropy loss ($\mathcal{L}_{ID}$) and  cross-camera feature align loss ($\mathcal{L}_{CCFA}$) are employed for representation learning.}
			% the cross-camera stable feature generation  (CCSFG) for the ISCS person Re-ID. The encoder $E$ extract the feature $\boldsymbol{f}$ from the input images.
		}
		\label{fig:framework}
	\end{figure}
	
	\subsection{Camera-Conditioned Stable Feature Generation}
	%{\color{red} Without cross-camera paired data, we need to generate.}
	
	We propose a new method, Camera-Conditioned Stable Feature Generation (CCSFG), for the ISCS person Re-ID problem. The model structure is illustrated in \cref{fig:framework}. It consists of two components, the person image encoder $E$ and the feature generator $G$. The common person Re-ID backbones, e.g., Resnet-50, can be used as $E$ and our novel $\sigma$-Reg.~CVAE as generator $G$.
	During training, $E$ and $G$ are jointly optimized for mutual improvements.
	On the one hand, the generator $G$ is trained with the feature extracted by $E$ as input, which is encoded by the loss $\mathcal{L}_{G|E}$.
	On the other hand, the features across different cameras $\{v_1,\dots,v_C\}$ are generated by $G$ for the training of $E$. This is modeled as the loss $\mathcal{L}_{E|G}$.
	The overall training objective of CCSFG thus becomes,
	\begin{equation}
		\mathop{\min}\limits_{G,E}\mathcal{L}_{(G,E)} = \mathcal{L}_{G|E} + \mathcal{L}_{E|G} \ ,
	\end{equation}
	where $E$ and $G$ are end-to-end optimized with $\mathcal{L}_{(G,E)}$.
	
	% As show in \cref{fig:framework}, we proposed the Camera-Conditioned Stable Feature Generation method (CCSFG), which is composed of Encoder $E$ and Generator $G$. $E$ is a  ReID model to produce feature vectors. $G$ is a generative model $\sigma$-Reg.~CVAE to generate the cross-camera samples in the feature space for model training. $E$ and $G$ are end-to-end trained under a unified framework. The training objective of CCSFG could be formulated as:
	% \begin{equation}
	%     \mathop{\min}\limits_{G,E}\mathcal{L}_{(G,E)}(x,y,c) = \mathcal{L}_{G|E}(x,y,c) + \mathcal{L}_{E|G}(x,y),
	% \end{equation}
	% where $\boldsymbol{x}$ denotes an input image, $y$ and $c$ denote the identity label and camera label, respectively.
	
	%As mentioned before, cross-camera paired data is crucial supervision to learn discriminative features for person Re-ID. In ISCS setting, due to lack of cross-camera paired data, we can't directly use the metric function to learn camera-unrelated feature directly. Therefore, we proposed Camera-Conditioned Stable Feature Generation pipeline (CCSFG) to generate cross-camera feature as shown in \cref{fig:framework}, which joints feature learning and feature generation. 
	%To relief the problem of cross-camera paired data in ISCS setting, we introduce the pipeline that can generate the cross-camera samples in the feature space for model training. Based on this pipeline, 
	
	To simplify the presentation of the training procedure, one training sample $(\boldsymbol{x},y,c)$ is considered by default. The extension to mini-batch is straightforward.
	The appearance feature of person image $\boldsymbol{x}$ is extracted by encoder $E$,
	% Given an input image $\boldsymbol{x}$, we can get the feature vector by $E$:
	\begin{equation} \label{eq:encoder}
		\boldsymbol{f} = E(\boldsymbol{x}),
	\end{equation}
	where $\boldsymbol{f} \in \mathbb{R}^{d}$.
	
	Given the image feature $\boldsymbol{f}$ under one camera, we expect the image features of the same person under other cameras to be generated. Therefore, the proposed generator G is built on the conditional variational autoencoder (CVAE) architecture for the convenience of introducing the side information, such as the identity $y$ and camera view $c$.
	
	% the generative architecture, conditional variational autoencoder (CVAE), is adopted for the convenience of introducing the side information, such as the identity $y$ and camera view $c$.
	% {\color{red} We expect to get a generator, which takes identity label and camera label as input, and generates a sample of a specific pedestrian under a certain camera view in the feature space. To this end, $\sigma$ Regularized Conditional Variational Autoencoder ($\sigma$-Reg.~CVAE) is proposed.} 

	%The estimation process of $\sigma$-Reg.~CVAE are illustrated the following.
	% \begin{figure*}[t]
	%     \centering
	%     \includegraphics[width=0.9\textwidth]{images/E_G.pdf}
	%     \caption{{\color{red}Overview of the proposed generator $\sigma$ Regularized Conditional Variational Autoencoder. (a) In the estimation stage, the generator $G$ takes the feature $\bar{f}$ identity label $y$ and camera label $y$ as input to maximize the conditional log-likelihood $\log P(\boldsymbol{\bar{f}}|c,y)$. (b) In the generation stage, the identity label $y$ and camera view $\{v_i\}_{i=1}^{C}$ is used as input to generate camera-conditioned features $\{\boldsymbol{g}_{v_i}\}_{i=1}^{C}$.}
	%     }
	%     \label{fig:E_G}
	% \end{figure*}
	
	\begin{figure}[t]
		\centering
		\includegraphics[width=0.42\textwidth]{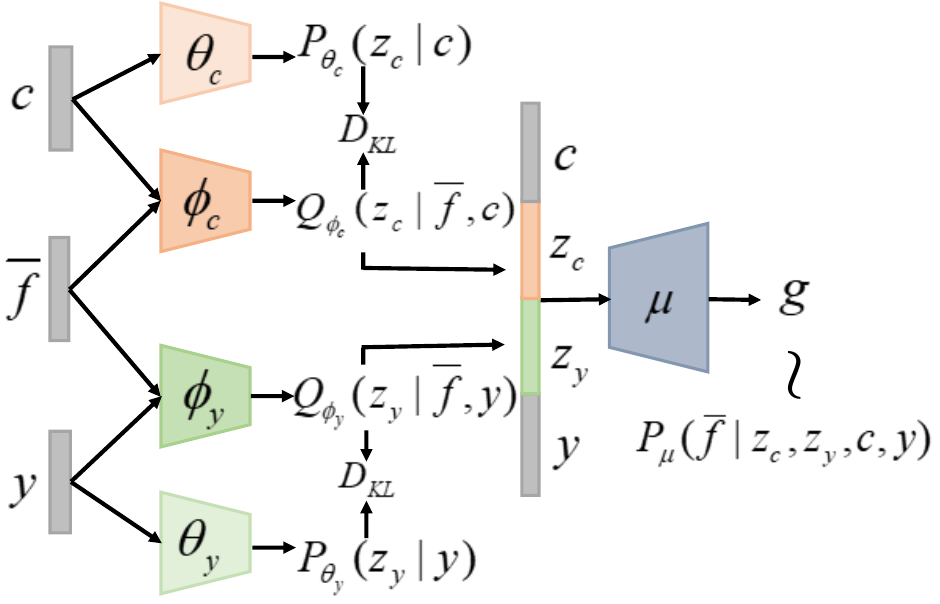}
		\caption{
			Illustration of the training (estimation) process of $\sigma$-Reg.~CVAE. The generated feature $g$ obeys the decoding distribution $P_\mu(\boldsymbol{\bar{f}}|z_c,z_y,c,y)$ given the feature $\bar{f}$, identity label $y$ and camera label $c$. 
			% The latent variables $\boldsymbol{z}_y$ and $\boldsymbol{z}_c$ are drawn form the posterior distributions. ...
			% The feature $\bar{f}$, identity label $y$ and camera label $c$ is used as input to maximize the conditional log-likelihood $\log P(\boldsymbol{\bar{f}}|c,y)$ ...
		}
		\label{fig:G_EST}
	\end{figure}
	
	\noindent\textbf{Training (Estimation) of $G$.}\quad 
	To learn the parameters of our generator $\sigma$-Reg.~CVAE, the loss $\mathcal{L}_{G|E}$ is used. It corresponds to the Estimation phase of $G$, as shown in \cref{fig:framework}.
	Specifically, image feature $\boldsymbol{f}$ is the input of $\sigma$-Reg.~CVAE and the normalized feature $\boldsymbol{\bar{f}}$ is obtained,
	\begin{equation}
		\boldsymbol{\bar{f}} = \operatorname{IFN}(\boldsymbol{f}),
	\end{equation}
	where $\operatorname{IFN}(\cdot)$ is a normalization function in $\sigma$-Reg.~CVAE. It plays the important role in the stable joint learning of CCSFG and will be discussed in greater details in \cref{sec:stable_train}.
	% input feature 
	% with detailed analysis and illustrations provided in \cref{sec:stable_train}.

	The direct learning objective of $\sigma$-Reg.~CVAE is maximizing the conditional log-likelihood $\log P(\boldsymbol{\bar{f}}|c,y)$, which is often intractable.
	Its variational lower bound is optimized instead by introducing latent variables~\cite{cvae}.
	Specially, $\boldsymbol{z}_y$ and $\boldsymbol{z}_c$ are the two latent variables introduced in the $\sigma$-Reg.~CVAE and correspond to the given identity condition $y$ and camera condition $c$ respectively. Their prior distributions, $P_{\theta_y}(\boldsymbol{z}_y|y)$ and $P_{\theta_c}(\boldsymbol{z}_c|c)$, are modeled with two prior networks $\theta_y$ and $\theta_c$.
	The two recognition networks $\phi_y$ and $\phi_c$ map $(\boldsymbol{\bar{f}},y)$ and $(\boldsymbol{\bar{f}},c)$ to their posterior distribution $Q_{\phi_y}(\boldsymbol{z}_y|\boldsymbol{\bar{f}},y)$ and $Q_{\phi_c}(\boldsymbol{z}_c|\boldsymbol{\bar{f}},c)$.
	% The latent variables $\boldsymbol{z}_y$ and $\boldsymbol{z}_c$  are obtained by sampling from $Q_{\phi_y}(\boldsymbol{z}_y|\boldsymbol{\bar{f}},y)$ and $Q_{\phi_c}(\boldsymbol{z}_c|\boldsymbol{\bar{f}},c)$.
	Moreover, the decoding distribution $P_{\mu}(\boldsymbol{\bar{f}} |\boldsymbol{z}_{c},\boldsymbol{z}_{y},c,y)$ is modeled by the decoder network $\mu$.
	Based on the previous sub-networks and distributions defined, an applicable learning objective of $\sigma$-Reg.~CVAE is,
	\begin{equation}
		\begin{aligned} \label{eq:my_CVAE}
			&\mathcal{L_{\text{EST}}}(\boldsymbol{\bar{f}}, y, c\ |\ \theta_y, \theta_c, \phi_y, \phi_c, \mu) = \\
			& \mathbb{E}_{Q_{\phi_y}(\boldsymbol{z}_y|\boldsymbol{\bar{f}},y)Q_{\phi_c}(\boldsymbol{z}_c|\boldsymbol{\bar{f}},c)}[-\log{P_{\mu}(\boldsymbol{\bar{f}} |\boldsymbol{z}_{c},\boldsymbol{z}_{y},c,y)}]   \\
			&+D_{KL}(Q_{\phi_y}(\boldsymbol{z}_y|\boldsymbol{\bar{f}},y)||P_{\theta_y}(\boldsymbol{z}_y|y))  \\
			&+D_{KL}(Q_{\phi_c}(\boldsymbol{z}_c|\boldsymbol{\bar{f}},c)||P_{\theta_c}(\boldsymbol{z}_c|c)),
		\end{aligned}
	\end{equation}
	where $D_{KL}$ denotes the Kullback-Leibler divergence. The construction of this loss is depicted in \cref{fig:G_EST}.

	The latent variables $\boldsymbol{z}_y$ and $\boldsymbol{z}_c$  are sampled from different distributions across stages.
	% at model estimation and feature generation. 
	During training $\boldsymbol{z}_c \sim Q_{\phi_c}(\boldsymbol{z}_c|\boldsymbol{\bar{f}},c) $ and $\boldsymbol{z}_y \sim Q_{\phi_y}(\boldsymbol{z}_y|\boldsymbol{\bar{f}},y)$ while testing $\boldsymbol{z}_c \sim P_{\theta_c}(\boldsymbol{z}_c|c)$ and $\boldsymbol{z}_y \sim P_{\theta_y}(\boldsymbol{z}_y|y)$.
	Such inconsistency can harm the quality of the generated feature samples.
	% More consistent $\boldsymbol{z}_y$ and $\boldsymbol{z}_c$ across stages results in better generative qualities.
	The Gaussian Stochastic Neural Network (GSNN)~\cite{cvae} method is exploited to alleviate this issue with a loss,
	\begin{equation}
		\begin{aligned}
			&\mathcal{L}_{GSNN}(\boldsymbol{\bar{f}}, y, c\ |\ \theta_y, \theta_c) = \\ &\mathbb{E}_{P_{\theta_y}(\boldsymbol{z}_y|y)P_{\theta_c}(\boldsymbol{z}_c|c)}[-\log{P_{\mu}(\boldsymbol{\bar{f}} |\boldsymbol{z}_{c},\boldsymbol{z}_{y},c,y)}].
		\end{aligned}
	\end{equation}
	
	% In the generation phase, the recognition networks are discarded, and 
	% $\boldsymbol{z}_c$ and $\boldsymbol{z}_y$ are drawn from the prior distribution $P_{\theta_c}(\boldsymbol{z}_c|c)$ and $P_{\theta_y}(\boldsymbol{z}_y|y)$. 
	% It introduces inconsistency between the training and generation stage and affects the quality of feature generation. To remedy this, we use the Gaussian Stochastic Neural Network method (GSNN)\cite{cvae}. The objective function of GSNN can be written as:
	% \begin{equation}
	%     \begin{aligned}
	%     &\mathcal{L}_{GSNN}(\boldsymbol{\bar{f}}, y, c\ |\ \theta_y, \theta_c) = \\ &\mathbb{E}_{P_{\theta_y}(\boldsymbol{z}_y|y)P_{\theta_c}(\boldsymbol{z}_c|c)}[-\log{P_{\mu}(\boldsymbol{\bar{f}} |\boldsymbol{z}_{c},\boldsymbol{z}_{y},c,y)}].
	%     \end{aligned}
	% \end{equation}
	
	The overall objective for the estimation of the generator $\sigma$-Reg.~CVAE is,
	% The overall loss for the parameter estimation of our generator $\sigma$-Reg.~CVAE is,
	\begin{equation} \label{eq:hybrid}
		\min_{\theta_y, \theta_c, \phi_y, \phi_c, \mu} \mathcal{L}_{G|E} = \alpha\mathcal{L_{\text{EST}}}+(1-\alpha)\mathcal{L}_{GSNN},
	\end{equation}
	with $\alpha$ as the balancing hyper-parameter. 
	
	%In the generation phase, the input of $\sigma$-Reg.~CVAE is identity label $y$ and camera label $\{v_i\}_{i=1}^{C}$. Specially, given the $y$ and $v_i$, we can get the prior distribution of $p(\boldsymbol{z}_y|y)$ and $P(\boldsymbol{z_{v_i}}|v_i)$ by two prior networks. Then, we sample $\boldsymbol{z}_y \sim P(\boldsymbol{z}_y|y)$ and $\boldsymbol{z_{v_i}} \sim P(\boldsymbol{z_{v_i}}|v_i)$ and feed $(v_i,p,\boldsymbol{z_{v_i}},\boldsymbol{z}_y)$ to the decoder network. Finally, we can get the generated feature $\boldsymbol{g}_{v_i}$ of $y$ under camera view $v_i$. For the feature $\boldsymbol{\bar{f}}$, we can obtain $c$ views of features $\{\boldsymbol{g}_{v_i}\}_{i=1}^{C}$.
	\begin{figure}[t]
		\centering
		\includegraphics[width=0.4\textwidth]{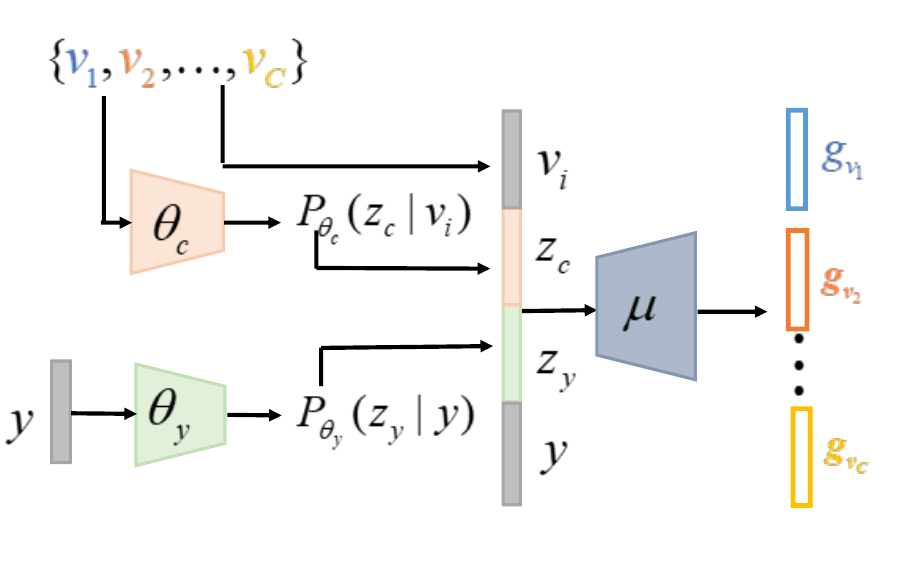}
		\caption{Illustration of the generation process of $\sigma$-Reg.~CVAE. 
			Conditioning on the identity label $y$ and different camera views $\{v_i\}_{i=1}^{C}$ as inputs, the corresponding generated features $\{\boldsymbol{g}_{v_i}\}_{i=1}^{C}$ are obtained from the decoder network $\mu$.
			% The identity label $y$ and camera view $\{v_i\}_{i=1}^{C}$ is used as input to generate camera-conditioned features $\{\boldsymbol{g}_{v_i}\}_{i=1}^{C}$.
		}
		\label{fig:G_GEN}
	\end{figure}
	\noindent\textbf{Training of $E$.}\quad 
	The cross-camera images of the same person play the central role in training image encoder $E$ but not available under the ISCS setting.
	% The person images captured by different cameras play the central role in training the discriminative camera-view invariant feature encoder $E$.
	% However, such cross-camera images of the same person do not exist under the ISCS setting.
	The feature samples of a person  under different camera views are compensated from our generator $G$, $\sigma$-Reg.~CVAE, for the training of encoder $E$.
	Therefore, $\mathcal{L}_{E|G}$ is used to indicate the overall training loss of $E$.
	
	% To extract the discriminative and camera-view invariant features of person images, 
	% To extract the camera-unrelated and ID-discriminative feature, we apply $\mathcal{L}_{E|G}$ to update the parameters of encoder $E$. $\mathcal{L}_{E|G}$ is composed of three losses three loss functions as follows.
	
	To obtain the synthesized features, the proposed $\sigma$-Reg.~CVAE is functioned under the Generation mode, as illustrated in \cref{fig:G_GEN}.
% 	{\color{red} We use the person identity $y$ of $\bar{\boldsymbol{f}}$, the camera views $c = v_i$, $v_i\in \{v_1,...,v_C\}$ and the latent variables $\boldsymbol{z}_y \sim P_{\theta_y}(\boldsymbol{z}_y|y)$ and $\boldsymbol{z}_{c} \sim P_{\theta_c}(\boldsymbol{z}_{c}|v_i)$ as the input of the decoder network $\mu$ to generate camera-conditioned features $g$,}
	{\color{black} With an input feature $\boldsymbol{\bar{f}}$, its person identity label $y$, the camera views $v_i$, $v_i\in \{v_1,...,v_C\}$ and the latent variables $\boldsymbol{z}_y \sim P_{\theta_y}(\boldsymbol{z}_y|y)$ and $\boldsymbol{z}_{c} \sim P_{\theta_c}(\boldsymbol{z}_{c}|v_i)$ given, camera-conditioned features $\boldsymbol{g}$ can be generated from the decoder network $\mu$ of $G$, }
% 	With an input feature $\bar{\boldsymbol{f}}$ and its person identity $y$, the camera views $c = v_i$, $v_i\in \{v_1,...,v_C\}$ specified and
% 	the latent variables sampled from the priors $\boldsymbol{z}_y \sim P_{\theta_y}(\boldsymbol{z}_y|y)$ and $\boldsymbol{z}_{c} \sim P_{\theta_c}(\boldsymbol{z}_{c}|v_i)$,
% % 	and the $y$ and camera $v_i$ specified,
% 	the generated feature can be computed from the decoder network $\mu$ of $G$,
	\begin{equation}\label{eq:g_mu}
		\boldsymbol{g}_{v_i} = \mu(\boldsymbol{z}_{c}, \boldsymbol{z}_y, v_i, y).
	\end{equation}
	Different $\{\boldsymbol{g}_{v_i}\}_{i=1}^{C}$ are generated by keeping the identity label $y$ the same and traversing over cameras $v_i$.
% 	where $y$ is the identity label of input $\bar{\boldsymbol{f}}$ and thus not specified. More importantly, 
	Therefore, $\boldsymbol{\bar{f}}$ and its corresponding generated features $\{\boldsymbol{g}_{v_i}\}_{i=1}^{C}$ form the cross-camera samples of the same person. 
% 	This generation procedure is illustrated in \cref{fig:G_GEN}.
	
	Different discriminative loss can then be applied on $\boldsymbol{\bar{f}}$ and $\{\boldsymbol{g}_{v_i}\}_{i=1}^{C}$ for the learning of encoder $E$.
	On the one hand, conditioning on the identity label $y$ and camera views $v_i$, $i = 1,...,C$, the generated features $\{\boldsymbol{g}_{v_i}\}_{i=1}^{C}$ from the generator $\sigma$-Reg.~CVAE are ID discriminative and camera view specified. However, an ideal encoder $E$ should extract the discriminative and view-invariant feature $\boldsymbol{\bar{f}}$ from a person image.
	To achieve this goal, the averaged distance between an image feature $\boldsymbol{\bar{f}}$ and the corresponding  $\{\boldsymbol{g}_{v_i}\}_{i=1}^{C}$ should be minimized. By pulling $\boldsymbol{\bar{f}}$ towards different $\boldsymbol{g}_{v_i}$ can not only preserves their id distinctive information but also eliminating the camera-view dependent information in $\boldsymbol{\bar{f}}$.
	We propose a novel Cross-Camera Feature Align (CCFA) loss for the purpose described above,
	\begin{equation} \label{eq:gsl}
		\mathcal{L}_{CCFA}(\boldsymbol{\bar{f}} \ | \ G) = \frac{1}{C}\sum_{i=1}^{C} ||\boldsymbol{\boldsymbol{\bar{f}}}- \boldsymbol{g}_{v_i}||^2,
	\end{equation}
	where $||\cdot||$ denotes the feature norm. This loss is for the learning of encoder $E$ only.

	{\color{black}
	On the other hand, the cross-entropy loss $\mathcal{L}_{ID}$ is used,
	\begin{equation} \label{eq:id}
		\mathcal{L}_{ID}(y, \boldsymbol{\bar{f}} \ | \ G) = -\log(\boldsymbol{q}[y]),
	\end{equation}
	where $\boldsymbol{q}[y]$ denotes the identity predictions of $\boldsymbol{\bar{f}}$ on the ground-truth $y$.
	}
% 	We control the gradient-flow of generated features $\frac{\partial \mathcal{L}_{ID}}{\partial \boldsymbol{g}_{v_i}}$ back to the encoder $E$ during optimization to further enhance their impacts as cross-camera samples, as illustrated in \cref{fig:framework}.
	
	% For ID-discriminative feature learning, we use the cross-entropy loss $\mathcal{L}_{ID}$ as follows:
	% \begin{equation} \label{eq:id}
	%     \mathcal{L}_{ID} = -y\log(\boldsymbol{q}),
	% \end{equation}
	% where $y$ denotes the ground true label and $\boldsymbol{q}$ denotes the identity predictions of $\boldsymbol{\bar{f}}$. Since there are too many identities in the Re-ID datasets, we apply the label-smoothing method \cite{szegedy2016rethinking} to prevent our model from overfitting to training IDs.
	
	% For feature similarity learning, we use the MCNL loss \cite{SCT} as follows: 
	% \begin{equation}
	% \begin{aligned}
	%     \mathcal{L}_{MCNL} &= [m_1 + dist_{+,intra} - dist_{-, inter}]_{+}  \\
	%                       &+ [m_2 + dist_{-,inter}-dist_{-,intra}]_{+},
	% \end{aligned}
	% \end{equation}
	% where $[z]_{+} = max(z,0)$, $m_{1}$ and $m_{2}$ denote explicit margins, $dist_{+,intra}$ denotes the distance between the feature $\boldsymbol{f}$ and intra-camera hard positive pair, $dist_{-, inter}$ denotes the distance between the feature $\boldsymbol{f}$  and inter-camera hard negative pair, $dist_{-,intra}$ denotes the distance between the feature $\boldsymbol{f}$  and inter-camera hard negative pair. 
	Moreover, the MCNL loss~\cite{SCT}, denoted as $\mathcal{L}_{MCNL}$, is also exploited for the feature similarity learning on the extracted person image features by $E$.
	By aggregating the training losses for $E$, the overall loss $\mathcal{L}_{E|G}$ is,
	\begin{equation} \label{eq:obj_E}
		\mathcal{L}_{E|G} =\lambda_{1}\mathcal{L}_{CCFA} +\lambda_{2}\mathcal{L}_{ID}  + \lambda_{3}\mathcal{L}_{MCNL},  
	\end{equation}
	where $\lambda_{1}$, $\lambda_{2}$ and $\lambda_{3}$ are balancing hype-parameters.
	
	The encoder $E$ and our generator $G$, $\sigma$-Reg.~CVAE, are end-to-end optimized with the following loss,
	%\noindent\textbf{Joint optimization.}
	% We jointly train the encoder and generator to optimize the total objective, which is a weighted sun of the following losses:
	\begin{equation}
		\begin{aligned} 
			\mathcal{L}_{(G,E)} &=  \alpha\mathcal{L_{\text{EST}}}+(1-\alpha)\mathcal{L}_{GSNN} \\ 
			&+\lambda_{1}\mathcal{L}_{CCFA} +\lambda_{2}\mathcal{L}_{ID} + \lambda_{3}\mathcal{L}_{MCNL}.
		\end{aligned}
	\end{equation}
	During testing, the person appearance feature are extracted by the learned encoder $E$ for retrieving.
	% To further improve the performance of the model, we also implement local branch \cite{ccfp}. During testing, the generative model $\sigma$-Reg.~CVAE is discarded and features $\boldsymbol{\bar{f}}$ are used for retrieving. 
	
	% \subsection{$\sigma$-Regularized Conditional Variational Autoencoder}
	\subsection{The stability in training $\boldsymbol{G}$}
	\label{sec:stable_train}
	
\begin{figure}[t]
		\centering
		\includegraphics[width=0.49\textwidth]{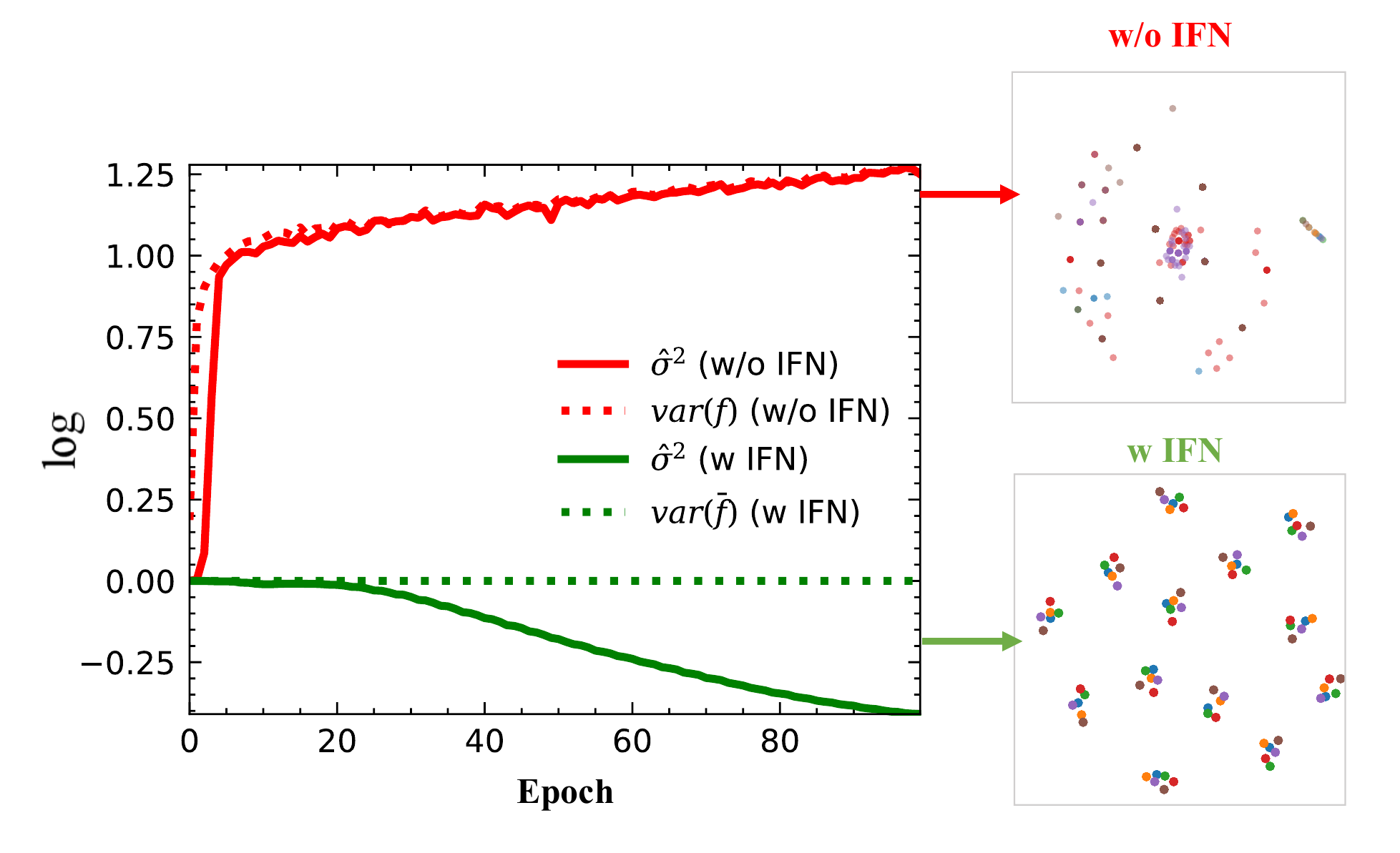}
		\caption{
			The curves on the left are the values of $\hat{\sigma}^2$ and $var(\boldsymbol{f})$ of the generative models with and without IFN during training on Market-SCT.
			% The values of $\sigma^2$ and $var(\boldsymbol{f})$ during model training on Market-SCT. 
			The feature map, w/o IFN, on the top-right corresponds to the \emph{failure} training cases.
			The feature map, w IFN, at the bottom-right corresponds to the \emph{successful} joint learning.
			Each dot in a feature map is a generated feature of a person and the colors indicate the different conditioning camera views.
			% {\color{red} Red indicates that CCSFG is trained without input feature normalization (IFN). Green indicates that CCSFG is trained with IFN. In the two figures on the right, different colors indicate the generative features under different camera views.} 
			%Same as the analysis in the main text, $var(f)$ will increase very quickly at the beginning of training, and $\sigma^2$ will increase accordingly under the influence of the \cref{eq:sigma_decomp}. It makes the generator $G$ appear posterior collapse and generate meaningful features difficultly.}
		}
		\label{fig:sigma_curve}
	\end{figure}

	%Preliminary:
	In the proposed CCSFG, the encoder $E$ and generator $G$ are jointly learned. The image feature $\boldsymbol{f}$ is extracted by $E$ and used as the input for training $G$.
	However, $\boldsymbol{f}$ will be drastically changed across training steps as its encoder $E$ is also training.
	Based on such inputs, the optimization on generator $G$ can easily fail and ruin the whole learning procedure.
	As illustrated in \cref{fig:sigma_curve}, the generated features from the failure training are compared with the successful one. The meaningless scattered features are produced by the failure case (top-right) while the more ideal features with the clear and meaningful clusters (IDs) are from the stable training (bottom-right).
	% {\color{red} As shown in \cref{fig:sigma_curve}, $G$ generates some meaningless feature when $f$ is used as input directly. On the contrary, when $\bar{f}$ is used as the input to $G$, the generative features of different identity pedestrians form different clusters. }
	
	To achieve the successful joint learning of $G$ and $E$, the proposed $\sigma$-Reg.~CVAE is exploited as generator $G$. Its $\operatorname{IFN(\cdot)}$ module normalizes the image feature $\boldsymbol{f}$ to $\boldsymbol{\bar{f}}$ as input for feature generation and plays the key role in stabilizing the whole training process.
	% is the key component to stabilize its training.
	% a novel generator $\sigma$-Reg.~CVAE is exploited. The feature normalization function $\operatorname{IFN(\cdot)}$ is the key component to stabilize its training.
	Without $\operatorname{IFN(\cdot)}$, the $\sigma$-Reg.~CVAE degenerates to a conventional CVAE using the image feature $\boldsymbol{f}$ as input.
	% conditioned on identity $y$ and camera view $c$.
	Joint learning of image encoder $E$ and the conventional CVAE as $G$ will end up with posterior collapse.
	The explanations on them are provided as follows.
	% The theory of posterior collapse~\cite{Rybkin2021SimpleAE, takida2021preventing, CollapseICML2020} can be used to explain the failure of training a conventional CVAE under joint learning framework.
	
	Considering the conventional CVAE (w/o IFN) is used as the generator $G$.
	Its learning objective follows \cref{eq:my_CVAE} as $\mathcal{L_{\text{EST}}}(\boldsymbol{f}, y, c\ |\ \theta_y, \theta_c, \phi_y, \phi_c, \mu)$ with $\boldsymbol{f}$ rather than $\boldsymbol{\bar{f}}$ as input. Following~\cite{Rybkin2021SimpleAE, takida2021preventing, CollapseICML2020}, this loss can be rewritten as,
	\begin{equation}
		\begin{aligned} \label{eq:sigma_CVAE}
			\mathcal{L}_{\text{EST}} =& 
			\frac{d}{\sigma^2} ||\boldsymbol{f}-\boldsymbol{g}||^2 +  \frac{d}{2} \ln \sigma^2 \\
			&+D_{KL}(Q_{\phi_y}(\boldsymbol{z}_y|\boldsymbol{f},y)||P_{\theta_y}(\boldsymbol{z}_y|y))  \\
			&+D_{KL}(Q_{\phi_c}(\boldsymbol{z}_c|\boldsymbol{f},c)||P_{\theta_c}(\boldsymbol{z}_c|c)),
		\end{aligned}
	\end{equation}
	by replacing $\mu(\boldsymbol{z}_{c},\boldsymbol{z}_{y},c,y)$ with $\boldsymbol{g}$ as in \cref{eq:g_mu} and assuming the decoding distribution $P_{\mu}(\boldsymbol{f}|\boldsymbol{z}_{c},\boldsymbol{z}_{y},c,y)$ as an Isotropic Gaussian distribution,
% 	in \cref{eq:my_CVAE}
	\begin{align}\label{eq:decoder_dis}
		P_{\mu}(\boldsymbol{f}|\boldsymbol{z}_{c},\boldsymbol{z}_{y},c,y) = \mathcal{N}(\boldsymbol{g},\sigma^{2}I),
	\end{align}
	where $d$ in \cref{eq:sigma_CVAE} is the feature dimension of $\boldsymbol{f}$.
	% $\sigma^{2}$ is variance of the generated features and $d$ in \cref{eq:sigma_CVAE} is the feature dimension of $\boldsymbol{f}$. 
	\cref{eq:decoder_dis} assumes the input feature $\boldsymbol{f}$ obeys the Isotropic Gaussian distribution with mean as the generated feature $\boldsymbol{g}$ and a variance value as $\sigma^2$. With the feature pairs $(\boldsymbol{f}, \boldsymbol{g})$ available, $\sigma^2$ is estimated via,
	\begin{align}\label{eq:est_sigma}
		%\hat{\sigma}^2 & = \frac{1}{Md} \sum_{n=1}^{M} ||\boldsymbol{f}_n - \boldsymbol{g}_n||^{2}.
		\hat{\sigma}^2 = \frac{1}{d}\mathbb{E}(||\boldsymbol{f}-\boldsymbol{g}||^2).
	\end{align}
	Moreover, the input feature $\boldsymbol{f}$ has its own variance $var(\boldsymbol{f})$ defined as,
	\begin{align}\label{eq:var_f}
		var(f) = \frac{1}{d} \mathbb{E}(||\boldsymbol{f}-\mathbb{E}(\boldsymbol{f})||^2).
	\end{align}
% 	{\color{red} }
	With the alignment loss $\mathcal{L}_{CCFA}$ (\cref{eq:gsl}) for the training of $E$ the reconstruction term in $\mathcal{L}_{\text{EST}}$ (\cref{eq:sigma_CVAE}) for the training of $G$, the intrinsic characteristics of input $\boldsymbol{f}$, e.g., $\mathbb{E}(\boldsymbol{f})$, can be captured by the decoder network $\mu$ (for $\boldsymbol{g}$ generation as in \cref{eq:g_mu}) in CVAE, i.e., $\boldsymbol{g} \approx \mathbb{E}(\boldsymbol{f})$, and thus,
	\begin{align}\label{eq:sig_appx_var_f}
		%\hat{\sigma}^2 \approx \sigma^2 \approx var(\boldsymbol{f}).
		\sigma^2 \approx var(\boldsymbol{f}).
	\end{align}
	
\begin{figure}[t]
		\centering
		\includegraphics[width=0.47\textwidth]{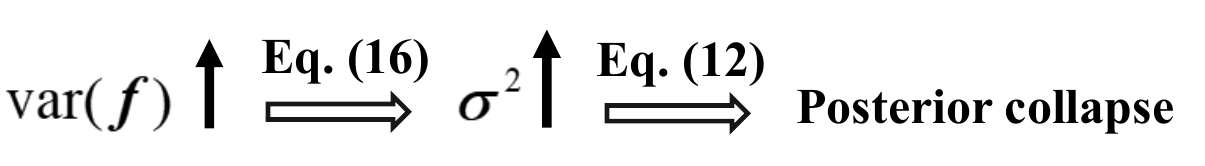}
		\caption{The impacts of huge $var(\boldsymbol{f})$ on the training of CVAE.
			%and $\sigma^2$ on each other. {\color{red} $var(f)$ and $\sigma^2$ are the variance of the feature $\bar{f}$ and decoding distribution.}
			%At the initial stage of training, the training of encoder leads to huge $var(\boldsymbol{f})$. As the training progresses, $var(\boldsymbol{f})$ and  $\sigma^2$ keep growing under \cref{eq:obj_E} and \cref{eq:sigma} effects.}
			% Two patterns, without $\&$ with $\operatorname{IFN(\cdot)}$, are illustrated.
		}
		\label{fig:Var(f)_sigma}
	\end{figure}

	In the joint learning procedure, the training of encoder $E$ leads to the rapid changes in image features and thus huge $var(\boldsymbol{f})$ occurs.
	Without normalization on the input $\boldsymbol{f}$, CVAE captures such variance and results in large $\sigma^2$ as in \cref{eq:sig_appx_var_f}. A concrete example of the joint learning of $E$ and CVAE on a ISCS dataset is shown in \cref{fig:sigma_curve} (left), where $\sigma^2$ is approximated by $\hat{\sigma}^2$. As the red curves shown, the values of both $var(\boldsymbol{f})$ and $\hat{\sigma}^2$ rise drastically as expected.
	However, large value of $\sigma^2$ prevents the CVAE to learn from the its input as the weight $\frac{d}{\sigma^2}$ on reconstruction term $||\boldsymbol{f}-\boldsymbol{g}||^2$ in $\mathcal{L}_{\text{EST}}$ \cref{eq:sigma_CVAE} becomes relatively small. This is known as the Posterior Collapse~\cite{Rybkin2021SimpleAE, takida2021preventing, CollapseICML2020} in training VAEs. The analysis above is depicted in \cref{fig:Var(f)_sigma}.
	The failure in training $G$ also ruins the training of $E$. 
	
	The $\operatorname{IFN(\cdot)}$ exploited by our $\sigma$-Reg.~CVAE is a simple statistical standardization technique as,
% 	The $\operatorname{IFN(\cdot)}$ function is exploited by our $\sigma$-Reg.~CVAE to normalize the input feature $\boldsymbol{f}$ from encoder. It is a simple statistical standardization technique as,
	\begin{equation}
		\boldsymbol{\boldsymbol{\bar{f}}} = \operatorname{IFN(\boldsymbol{f})} = \frac{\boldsymbol{f}-\mathbb{E}(\boldsymbol{f})}{\sqrt{var(\boldsymbol{f})+\epsilon}},
	\end{equation} 
	where $\epsilon$ is a small value.
% 	$\mathbb{E}(\boldsymbol{f})$ and $var(\boldsymbol{f})$ denote the mean and variance of the feature, and $\epsilon$ is a small value.
	It puts a hard constraint $var(\boldsymbol{\bar{f}}) = 1$ on the input $\boldsymbol{\bar{f}}$ to CVAE and eliminates the impact of drastic changes on inputs to the generator.
	% , especially at the initial training stage.
	From \cref{eq:sig_appx_var_f}, the value of $\sigma^2$ is thus regularized by the introduction of $\operatorname{IFN}$.
	Therefore, our proposed generator is called $\sigma$-Reg.~CVAE to highlight such a mechanism.
	The values of $var(\boldsymbol{\bar{f}})$ and $\hat{\sigma}^2$ with $\operatorname{IFN}$ are the green curves in \cref{fig:sigma_curve}. $var(\boldsymbol{\bar{f}})$ fixed at 1 because of $\operatorname{IFN(\cdot)}$ applied. $\hat{\sigma}^2 \approx 1$ at first according to \cref{eq:sig_appx_var_f}. However, $\hat{\sigma}^2$ is the estimation rather than $\sigma^2$ itself. As shown in \cref{eq:est_sigma}, $\hat{\sigma}^2$ also reflects the reconstruction loss in $\mathcal{L}_{\text{EST}}$ \cref{eq:sigma_CVAE}, which is gradually decreasing during training, as the green solid curve behaves.

	\section{Experiments}
	
	% \subsection{Dataset and setting}
	
	\noindent\textbf{Datasets.}\quad
	To evaluate and compare different methods under the ISCS person Re-ID settings, two benchmark datasets~\cite{SCT, ccfp}, i.e., Market-SCT and  MSMT-SCT, are exploited.
	Such datasets are built on the source ones, Market-1501\cite{market} and MSMT17\cite{msmt}, via only keeping the images of each person from one single camera for training. 
	With no cross-camera person images and fewer training samples, the datasets under the ISCS setting are much more challenging than the source ones.
% 	, as shown in \cref{tab:dataset}.
	Note that MSMT17 is a challenging dataset with the person images collected from different time periods and across largely varied scenes. It contains much more camera views, 15, than its counterparts with 6 and 8 only. Therefore, the MSMT-SCT can better stimulates the ISCS person Re-ID scenario.
	With the testing data unchanged, conventional person Re-ID evaluation metrics, Cumulative Matching Characteristic (CMC) and Mean Average Precision (mAP), are reported.
	
	% We evaluate our methods on three benchmark person Re-ID datasets, Market-SCT, Duke-SCT and MSMT-SCT, which are subset of Market-1501\cite{market}, DukeMTMC\cite{duke} and MSMT17\cite{msmt}. Compared with the original datasets, we choose one camera for each pedestrian images as the training data and keep the original testing data\cite{SCT}. The detailed information of the datasets are shown in \cref{tab:dataset}.
	
	% \noindent\textbf{Evaluation Metrics.}\quad Following the common evaluation metrics for Re-ID, performance is evaluated by the Cumulative Matching Characteristic (CMC) and Mean Average Precision (mAP)
	
% 	\begin{table}[h]
% 		\centering
% 		\footnotesize
% 		\caption{Details of the datasets used in our experiments.}
% 		\begin{tabular}{l|c|c|c|c|c}
% 			\hline
% 			Dataset & Train &Train &Test & Test & X-camera \\ &IDs & Images &IDs & Images  & paired data   \\ \hline
% 			Market-1501 & 751 &12936 &750  &15913  &True \\ \hline
% 			Market-SCT & 751 &3561 &750  &15913  &False \\\hline
% % 			DukeMTMC & 702 &16522 &1110  &17661  &True \\\hline
% % 			Duke-SCT & 702 &5993 &1110  &17661  &False \\\hline
% 			MSMT17 & 1041 &32621 &3060 &93820 &True \\\hline
% 			MSMT-SCT & 1041 &6715 &3060  &93820  &False \\ \hline
% 		\end{tabular}
% 		\label{tab:dataset}
% 	\end{table}

	\noindent\textbf{Implementation Details.}\quad
	Our image feature encoder $E$ is the ImageNet pre-trained Resnet-50, following existing work~\cite{SCT, ccfp} for fair comparison.
% 	{\color{red}Follow the CCFP\cite{ccfp}, we use local branch to further improve the performance of the model.}
	We also adopt the architecture with local branches as in \cite{ccfp}.
	The mini-batch size is set to $128$ with image data augmentation~\cite{He2020FastReIDAP}. 
% 	For model optimization, we use the 
	Adam optimizer is used with an initial learn rate $3.5 \times 10^{-4}$, which decays at the 100th and 000th epoch with a decay factor of 0.1, and a weight decay of $5 \times 10^{-4}$.
% 	{\color{red} We also applied L2 regularization technique to prevent the model from overfitting\cite{liu2019adaptive}. }
	The total number of training epochs is 500.
	The hyper-parameters $\alpha$, $\lambda_1$, $\lambda_2$ and $\lambda_3$ are set to 0.2, 0.5, 4, 1, respectively.
	All experiments can be run on an NVIDIA 2080Ti GPU.

	% we also implement local branch, use adaptive loss
	
	% For a fair comparison, we use ResNet-50 \cite{he2016deep} which is pre-trained on ImageNet \cite{Krizhevsky2012ImageNetCW} as our network encoder. The mini-batch size is set to 128. For each mini-batch, we randomly sample images from 2 cameras \cite{ccfp}. The data augmentation strategy followed the implementation in fast-reid \cite{He2020FastReIDAP}. For updating our model, we use the Adam optimizer with an initial learn rate $3.5 \times 10^{-4}$, which decays at the 150th and 300th epoch with a decay factor of 0.1, and a weight decay of $5 \times 10^{-4}$. The total number of training epochs is set to 500. We also applied L2 regularization technique to prevent the model from overfitting\cite{liu2019adaptive}. The hype-parameters $\alpha$, $\lambda_1$, $\lambda_2$ and $\lambda_3$ are set to 0.2, 1, 1, 0.25, respectively. To speed up the training process and improve memory efficiency, we use the mixed-precision training throughout the process \cite{Micikevicius2018MixedPT}. All experiments are implemented in PyTorch and trained on an NVIDIA 2080Ti GPU. 
	
	\begin{table*}[t]
		\centering
		%\footnotesize
		\caption{
		The performance of different methods under the ISCS person Re-ID setting.
% 		Two benchmark datasets, Market-SCT and MSMT-SCT are used.
		$\dag$ denotes the re-ranking technique \cite{Re_Ranking} is used. 
% 		$\S$ represents methods training under the supervised setting.
		}
		\resizebox{0.75\textwidth}{!}{
			\begin{tabular}{c|cccc|cccc}
				\hline
				\multirow{2}{*}{Methods}  & \multicolumn{4}{c|}{MSMT-SCT} & \multicolumn{4}{c}{Market-SCT}  \\  \cline{2-9}
				& R-1    & R-5   & R-10  & mAP   & R-1    & R-5   & R-10  & mAP   \\  \hline
				
				PCB\cite{PCB}(ECCV'18)              &    -    &  -    &  -    &   -   & 43.5   &  -    &  -    & 23.5   \\
				Suh's method\cite{Suh}(ECCV'18)     &    -    &   -   &   -   &   -   & 48.0   &   -   &   -   & 27.3   \\
				MGN-ibn\cite{MGN}(ACMMM'18)         & 27.8   & 38.6  & 44.1  & 11.7  & 45.6   & 61.2  & 69.3  & 26.6   \\
				Bagtrick\cite{bagtrick}(CVPR'19)         & 20.4   & 31.0  & 37.2  & 9.8   & 54.0   & 71.3  & 78.4  & 34.0   \\
				AGW\cite{AGW}(TPAMI'21)             & 23.0   & 33.9  & 40.0  & 11.1  & 56.0   & 72.3  & 79.1  & 36.6   \\    %\cline{2-14}
				Center Loss\cite{CenterLoss}(ECCV16)       &   -     &   -  &   -   &  -    & 40.3   &   -   &    -  & 18.5  \\
				A-Softmax\cite{Softmax}(CVPR'17)        &   -     &   -    &   -   &  -    & 41.9   &  -   &   -    & 23.2 \\
				ArcFace\cite{ArcFace}(CVPR'19)          &   -     &   -    &   -   &   -   & 39.4   &  -    &   -    & 19.8   \\% \cline{2-14}
	
				SimSiam\cite{SimSiam}(CVPR'21)          & 2.8    & 5.9   & 8.4   & 1.2   & 36.2   & 51.9  & 59.1  & 18.0  \\   %\cline{2-14}
				MMD\cite{MMD}(ICML'15)              & 42.2   & 55.8  & 61.4  & 18.2  & 67.7   & 83.1  & 88.2  & 44.0  \\
				CORAL\cite{CORAL}(ECCV'16)            & 42.6   & 55.8  & 61.5  & 19.5  & 76.2   & 88.5  & 93.0  & 51.5  \\ %\cline{2-14}
				%Precise-ICS\cite{Precise}(WACV'21)      & 17.2   & 28.4  & 34.3  & 6.7   & 50.0   & 67.5  & 74.8  & 31.2  & 41.2  & 57.9  & 64.2  & 25.9 \\
				HHL\cite{HHL}(ECCV'18)              & 31.4   & 42.5  & 48.1  & 11.0  & 65.6   & 80.6  & 86.8  & 44.8   \\
				MCNL\cite{SCT}(AAAI'20)             & 26.6   & 40.0  & 46.4  & 10.0  & 67.0   & 82.8  & 87.9  & 41.6   \\
				CCFP\cite{ccfp}(ACMMM'21)            & 50.1   & 63.3  & 68.8  & 22.2  & 82.4   & 92.6  & 95.4  & 63.9  \\
				CCFP$^\dag$ \cite{ccfp}(ACMMM'21)  & {\color{blue}\textbf{54.9}}   & 65.0  & 69.5  & {\color{blue}\textbf{33.6}}  & 84.1   & 90.9  & 93.1  & {\color{blue}\textbf{78.2}}   \\ \cline{1-9}
				CCSFG(Ours)                   & 54.6   & {\color{blue}\textbf{67.7}}  & {\color{blue}\textbf{73.1}}  & 24.6  & {\color{blue}\textbf{84.9}}   & {\color{red}\textbf{94.3}}  & {\color{red}\textbf{96.2}}  & 67.7  \\
				CCSFG$^\dag$ (Ours)          &{\color{red}\textbf{ 61.2}}   & {\color{red}\textbf{71.1}}  & {\color{red}\textbf{75.1}}  & {\color{red}\textbf{37.8}}  & {\color{red}\textbf{87.1}}   & {\color{blue}\textbf{92.8}}  & {\color{blue}\textbf{95.0}}  & {\color{red}\textbf{82.6}}   \\   \hline
				% BagTricks$^\S$\cite{bagtrick}(CVPR'19)      & 63.4   &  - & -  &12.4   &94.5    &-   &  - &85.9   & 89.0  &  - & -  &78.6  \\    
				% AGW$^\S$\cite{AGW} (TPAMI'21)         & 68.3   &  - & -  &14.7   &95.1    & -  &  -  &87.8   & 89.0 &  - & -  &79.6  \\ \hline
		\end{tabular}}
		
		\label{tab:SOTA}
	\end{table*}

	\subsection{Results}
% 	\subsection{Comparison with State-of-the-Art Methods}
	
	%{\color{red} The reason why the model works well on the MSMT dataset. We should be focused on analysis and give a reason.}
	
	The proposed CCSFG is compared with different state-of-the-art methods. Besides the existing methods (CCFP\cite{ccfp}, MCNL\cite{SCT}) for the ISCS setting, other methods, such as the image generation (HHL\cite{HHL}), distribution alignment (MMD\cite{MMD}, CORAL\cite{CORAL}), self-supervised learning (SimSiam\cite{SimSiam}), metric learning (Center Loss\cite{CenterLoss}, A-Softmax\cite{Softmax}, ArcFace\cite{ArcFace}), and baselines (PCB\cite{PCB}, Suh's method\cite{Suh}, MGN-ibn\cite{MGN}, Bagtrick\cite{bagtrick}, AGW\cite{AGW}) are included. The results are shown in \cref{tab:SOTA}. 
	
	The proposed CCSFG achieves superior results to all its competitors. Clear margins can be observed between our CCSFG and the second place method CCFP\cite{ccfp} which is the state-of-the-art ISCS Re-ID model based on self-learning and feature alignment. 
	Comparing with CCFP, CCSFG achieves 6.3\% R-1 and 4.2\% mAP improvements on MSMT-SCT.
	Such improvements on Market-SCT are 3.0\% R-1 and 4.4\% mAP. 
	The ISCS setting of person Re-ID is challenging. Many existing methods fail to achieve the ideal performance on it.
	The image generation method HHL~\cite{HHL} can improve the baseline methods with the cross-camera images generated for training and achieve comparable performance to the ISCS method MCNL~\cite{SCT}. However, generating the person images with cross-camera view information captured is a challenging task.
	The distribution alignment methods, MMD\cite{MMD} and CORAL\cite{CORAL}, also achieve substantially good performance. They align the holistic feature distributions of different camera views. A feature alignment loss, $\mathcal{L}_{CCFA}$, is also used in CCSFG (\cref{eq:gsl}) to align the image feature and its generated features under different cameras.

	\begin{figure}[t]
		\centering
		\includegraphics[width=0.38\textwidth]{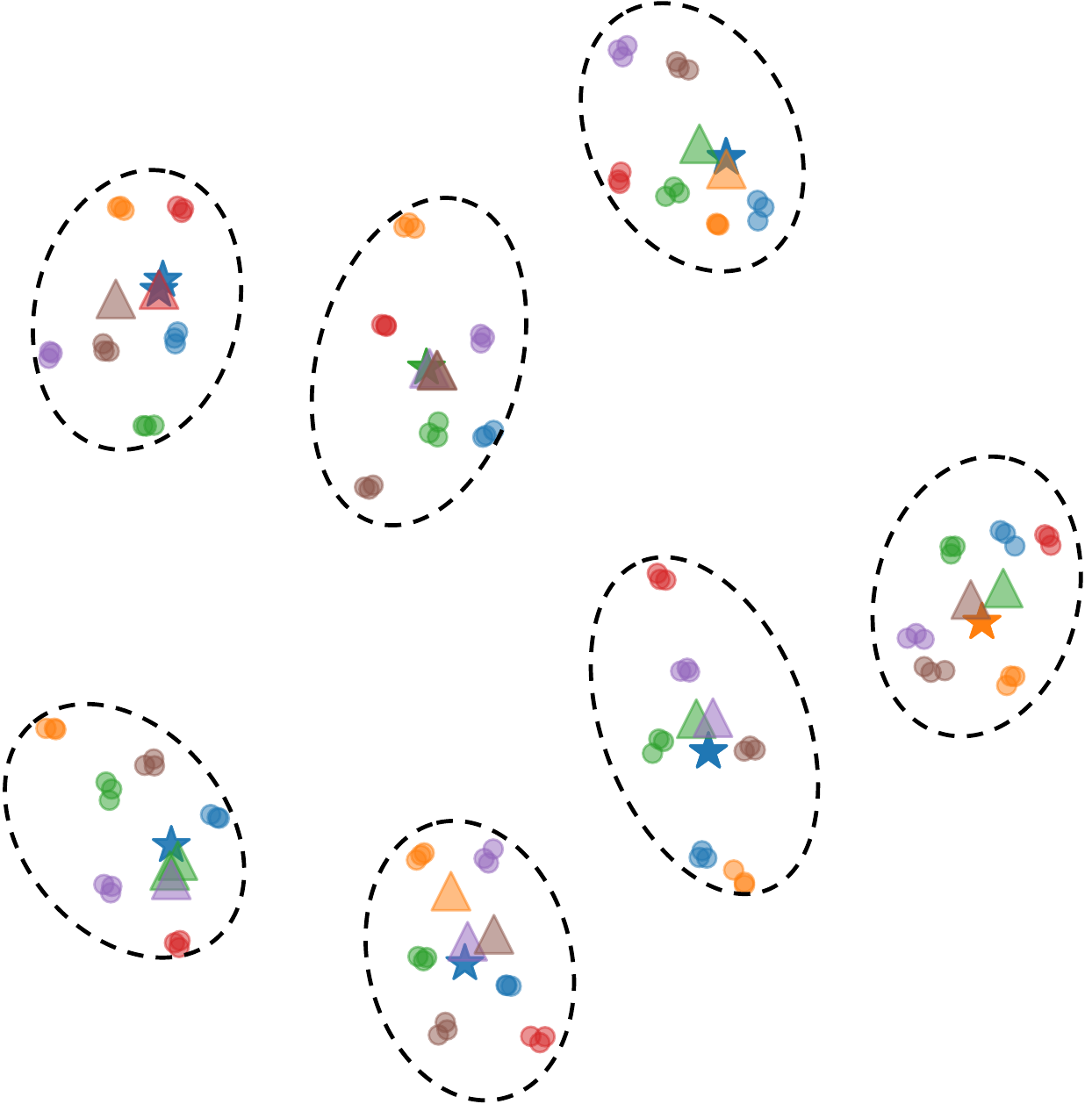}
		\caption{Visualization (t-SNE) of the training features produced by CCSFG on Market-1501. Dots are the generated features by $G$. Stars are the image features from $E$ of the images retained in Market-SCT. Triangles are the image features from $E$ of the deleted images of Market-1501 under the ISCS setting.
		Different colors indicate different cameras.
		The features in a oval belong to the same person.
% 		Different clusters represent features of different pedestrians. Different colors indicate features from different cameras. Dot, star and triangle denote generated features, existing features and missing cross-camera features in ISCS setting.
        }
		\label{fig:Fake_feature}
	\end{figure}

	\subsection{Detailed Analysis}
	
	In this subsection, we conducted the detail analysis on our CCSFG from different perspectives.
	
	\noindent\textbf{Visualization.}\quad 
	As shown in \cref{fig:Fake_feature}, meaningful features can be extracted and generated by the proposed CCSFG model.
	Firstly, obvious clusters are formed based on the person identities, which reflects that different features are discriminative.
	Secondly, the generated features $\boldsymbol{g}$s are id discriminative and view variant, as shown by the dense dots of different colors within a oval. Therefore, the generator $G$ can handle both the identity and camera view information provided by $y$ and $c$ to generate meaningful features $\boldsymbol{g}$s (dots).
	Thirdly, the features $\boldsymbol{f}$ extracted by $E$ on the training images under the ISCS setting are shown as stars. $\boldsymbol{f}$s are able to keep distances with the generated $\boldsymbol{g}$s under specified cameras.
	Moreover, the deleted cross-camera person images under the ISCS setting are fed into the trained encoder $E$ and their features are shown as triangles. The star and triangles of the same person are highly overlapped which indicates the discriminative and view-invariant person image features can be extracted by encoder $E$.
	Last but not least, such an ideal feature map demonstrates that the stable and effective joint training is achieved by CCSFG.

	\noindent\textbf{The analysis of model stability.}\quad
	The qualitative results of analyzing the stability of training with different generators are presented in \cref{sec:stable_train} along with the theoretical analysis. 
	The quantitative results are provided here to further evaluate the impact on joint learning with different generator models, as shown in \cref{tab:model_stable}. 
	Besides the comparisons with the vanilla CVAE \cite{cvae} that without any regularization on $\sigma^2$, the proposed $\sigma$-Reg.~CVAE is compared with the $\sigma$-CVAE \cite{takida2021preventing} which attempts to optimize the $\sigma^2$ value during training.
	However, neither the CVAE nor the $\sigma$-CVAE can stabilize the training procedure. Jointly learning the image feature encoder with such generators can even harm the performance.
	
% 	As mentioned before, the conventional CVAE training is unstable in our method. Since the variance of input feature $var(f)$ is gradually increasing during the training process, it will make the model appear posterior collapse and hardly generate meaningful features. It can be seen from \cref{tab:model_stable} that the performance of CVAE is lower than the baseline. This is because the generated features are almost noise which has negative impact on the training of encoder $E$. Even if other methods are used to control the value of $\sigma^2$ \cite{takida2021preventing}, it still can not achieve great results, because it ignores the influence on the feature distribution changes. It can be seen from the \cref{tab:model_stable} that  $sigma$-CVAE and CVAE achieved similar performance. Obviously, the key to stable training of the model is limit the variance of feature. When the IFN function is used, the distribution of feature have zero mean and unit variance, and $\sigma$-Reg.~CVAE can be successfully trained. As shown in \cref{tab:model_stable}, it brings 28.0\% Rank-1 and 14.6\% mAP increases in MSMT-SCT. Similar improvement can be observed in Market-SCT and Duke-SCT datasets. This suggests that using IFN can maintain the training stability of $\sigma$-Reg.~CVAE and help $E$ learn discriminative feature in ISCS setting.
	
	\begin{table}[t]
		\centering
		\footnotesize
		\caption{The joint learning stability with different generators. CCSFG is with our $\sigma$-Reg.~CVAE. Baseline is with encoder only.}
		\begin{tabular}{c|cc|cc}
			\hline
			\multirow{2}{*}{Methods}                    & \multicolumn{2}{c|}{MSMT-SCT} & \multicolumn{2}{c}{Market-SCT}  \\ \cline{2-5}
			& Rank-1          & mAP          & Rank-1         & mAP     \\  \hline
			Baseline             &26.6  &10.0   &67.0           &41.6      \\ 
			CVAE \cite{cvae}           & 12.1     & 5.2    & 58.1            & 37.7                     \\
			$\sigma$-CVAE \cite{takida2021preventing}      &11.2      & 4.1     &  59.3        &   38.6             \\
			CCSFG           &\textbf{54.6}   & \textbf{24.6}    & \textbf{84.9}  & \textbf{67.7}            \\ \hline
		\end{tabular}
		\label{tab:model_stable}
	\end{table}
	
	% \begin{table*}[]
	% \centering
	% % \scriptsize
	% \caption{Analysis on the model stability. }.
	% \begin{tabular}{c|cc|cc|cc}
	% \hline
	% \multirow{2}{*}{Methods}                    & \multicolumn{2}{c|}{Market-SCT} & \multicolumn{2}{c|}{Duke-SCT} & \multicolumn{2}{c}{MSMT-SCT} \\ \cline{2-7}
	%                           & Rank-1          & mAP          & Rank-1         & mAP     & Rank-1         & mAP     \\  \hline
	%         Baseline            & 75.7            & 51.5     & 70.1     & 53.1               &&\\
	%         CVAE                    &         59.3        &   38.6   &       60.1         &     40.1    &&    \\
	%         $\sigma$-CVAE                  & 58.1            & 37.7         &  59.6  &    39.1         &&\\
	%         CCSFG w/o p\&c                &           &      &        &          &&    \\
	%         CCSFG w/o c             &            &         &    &            &&\\
	%         CCSFG w/o p               &            &         &    &            &&\\ \hline
	% CCSFG              & \textbf{84.9}            & \textbf{67.7}         & \textbf{80.6}           & \textbf{64.5}         &&\\ \hline
	% \end{tabular}
	% \label{tab:model_stable}
	% \end{table*}

\begin{figure}[t]
		\centering
		\includegraphics[width=0.48\textwidth]{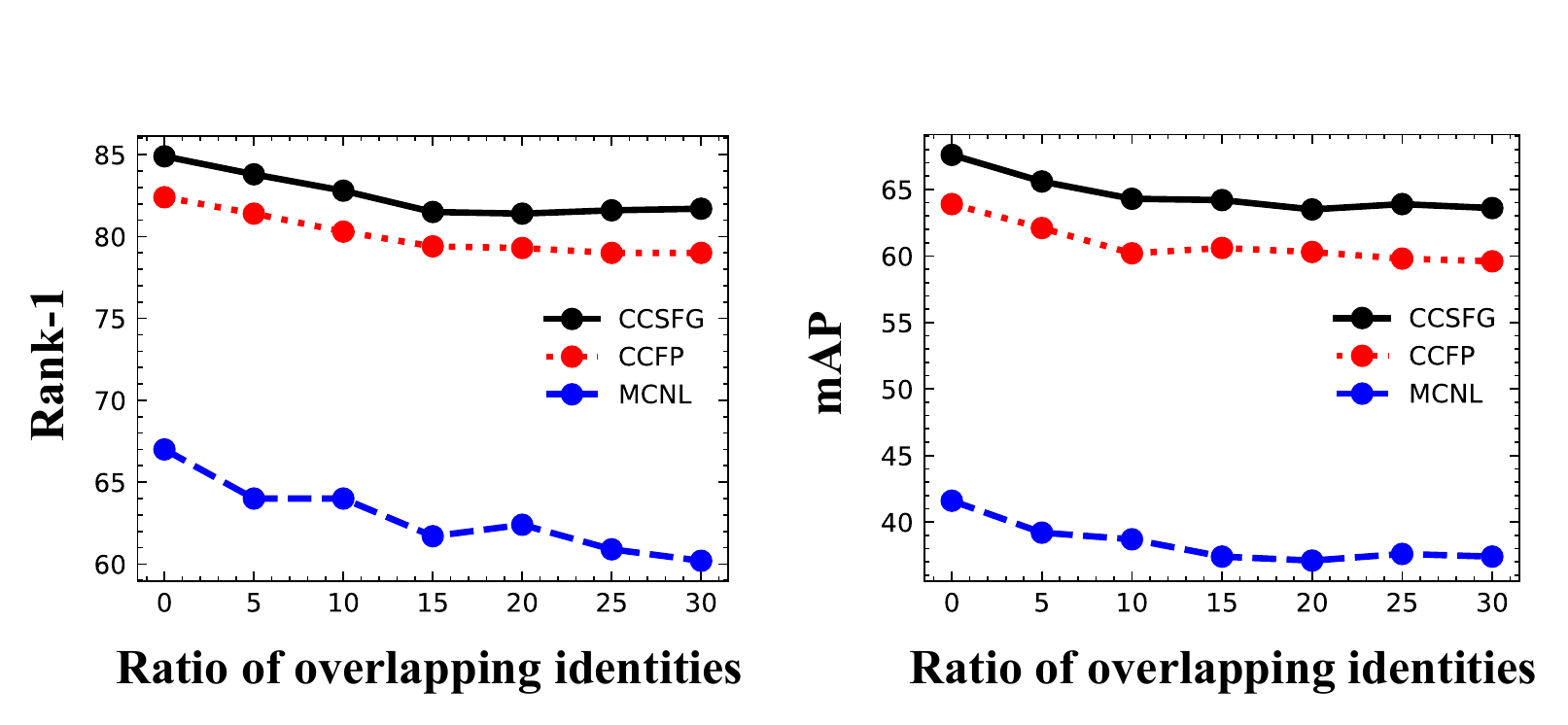}
		\caption{Analysis on the radios of overlapping identities on Market-1501. 
% 		(a) and (b) denote the results on the Market-SCT and Duke-SCT datasets, respectively.
		}
		\label{fig:Radio_overlapping}
	\end{figure}

	\noindent\textbf{The impact of the conditional variables.}\quad
	To verify the necessity of conditional variables $y$ and $c$ for feature generation, we conduct the ablation study on them, as shown in \cref{tab:model_condition}.
% 	we test it with different training setting. The results are shown in  \cref{tab:model_condition}.
	When both the identity label $y$ and camera label $c$ are not used in $G$ for feature generation,
% 	When CCSFG is implemented without identity labels $y$ and camera labels $c$, 
	our $\sigma$-Reg.~CVAE degenerates into a VAE-based model.
	Its generated feature $\boldsymbol{g}$ is not conditioned on camera and ID information, the corresponding training objective will be reducing the distance between $\boldsymbol{g}$ and the input feature $\boldsymbol{\bar{f}}$ only. As shown in the first row of \cref{tab:model_condition}, such the generator harms the person Re-ID performance.
% 	the camera and person information of generated feature $\boldsymbol{g}$ is unknown and we will reduce the distance between $\boldsymbol{g}$ and real feature $\boldsymbol{\bar{f}}$ as in \cref{eq:obj_E}.
% 	Compare with baseline, this methods drop the R-1 accuracy by 14.5\% on the MSMT-SCT dataset.
    Moreover, $G$ can incorporate either $y$ or $c$ only for feature generation. Substantial improvements can be obtained by considering more conditions, especially the identity label $y$.
    Since conditioning on $y$ can guarantee the discriminative power in the generated features and the encoder $E$ can benefit from them in the joint learning.
    Conditioning on both $y$ and $c$ in our generator, $\sigma$-Reg~CVAE, clearly boosts the performance.
% 	Similarly, When conditioned on $y$ or $c$ only, it is not able to control the corresponding condition of the generated feature and thus reducing the model performance. Compared with CCSFG, these two methods drop the Rank-1 accuracy by 32.0\% and 18.4\%.
	These results demonstrate the importance of both $y$ and $c$ for generating useful features in the joint learning of CCSFG.
	\noindent\textbf{Cross-camera identity overlap ratio.}\quad
	In the real-world surveillance application, fully non-overlapped persons across different cameras could be a strong assumption.
	Therefore, different ratios of cross-camera overlapping identities should be considered.
{\color{black} With the more cross-camera images of more persons existing (indicated by the higher ratio of overlapping IDs), the more training samples of same person are given distinctive ID labels. The obtained models are thus worse, as shown in \cref{fig:Radio_overlapping}.
However, Our CCSFG can withstand this challenge and stabilize at the SOTA level performance.}

% % 	As shown in \cref{fig:Radio_overlapping}, 
% 	the proposed CCSFG is more robust than the
% 	existing ISCS methods and consistently achieves the best performance under different radios of overlapping identities.

	\begin{table}[t]
		\centering
		\footnotesize
		\caption{
		%Ablation study on the conditional variables $y$ and $c$.
 		Analysis on the importance of condition variables to the generator. The condition variables $y$ and $c$ denote the identity label and camera label, respectively. $G$ is the $\sigma$-Reg.~CVAE.
		}
		\begin{tabular}{c|cc|cc}
			\hline
			\multirow{2}{*}{Methods}               & \multicolumn{2}{c|}{MSMT-SCT}     & \multicolumn{2}{c}{Market-SCT}   \\ \cline{2-5}
			& Rank-1          & mAP          & Rank-1         & mAP         \\  \hline
% 			Baseline             &26.6           &10.0   & 67.0           & 41.6     & 67.1     & 45.2   \\ 
			$G$ w/o $y$ \& $c$       &  12.4         &    6.3      &43.6     &  24.8             \\
			
			$G$ w/ $c$               &    22.6        &    12.3     &   49.3         &   31.6            \\
			$G$ w/ $y$             & 36.2  &15.3        & 77.1     & 54.3             \\
			G w/ $y$ \& $c$                &\textbf{54.6}   &\textbf{24.6}   & \textbf{84.9}            &\textbf{67.7}        \\ \hline
		\end{tabular}
		\label{tab:model_condition}
	\end{table}
	
% 	\noindent\textbf{The Hyper-parameter Analysis.} 
% 	The parameter $\alpha$ is significant to the training of $\sigma$-Reg.~CVAE, so we evaluate the value of $\alpha$ influence on the model results. We consider $\alpha$ from 0 to 1 with an interval of 0.2. The corresponding Rank-1 results are shown in \cref{fig:Ab_alpha}. CCSFG achieves the best Rank-1 accuracy when $\alpha = 0.2$. As mentioned before, $\alpha$ is used to reduce discrepancy between $\sigma$-Reg.~CVAE training and generation. When the value of $\alpha$ is closer to zero,  $\sigma$-Reg.~CVAE can generate more realistic features, but it will make training more difficult \cite{cvae}. Therefore, $\alpha=0.2$ is an appropriate value for CCSFG in ISCS setting.
% 	\begin{figure}[t]
% 		\centering
% 		\includegraphics[width=0.45\textwidth]{images/Alpha.pdf}
% 		\caption{Evaluation of parameter $\alpha$ in \cref{eq:hybrid}.}
% 		\label{fig:Ab_alpha}
% 	\end{figure}

	% $y$
	% \begin{figure}[t]
	%     \centering
	%     %\includegraphics{}
	%     \caption{p}
	%     \label{fig:Ab_p}
	% \end{figure}
	
	\section{Conclusion}
	
	%defect: 1. pose information 2. GAN
	
	In this paper, we focus on handling the challenging ISolated Camera Supervised (ISIC) person Re-ID problem where the cross-camera image pairs are not available for model training.
	To compensate the missing cross-camera data pairs, a novel pipeline based on feature generation is introduced.
	Following this pipeline, we propose the camera-conditioned stable feature generation (CCSFG), the first method to synthesize the cross-camera feature samples and end up with the joint learning between image encoder $E$ and feature generator $G$.
% 	The stability of $G$ becomes the main concern for joint learning.
    A novel generative model, $\sigma$-Reg.~CVAE, is then proposed as $G$ to achieve stable joint learning. The effectiveness of CCSFG is demonstrated by theoretical analysis and experimental results.
% 	With theoretical analysis and experimental demonstration, we find a simple yet effective solution to this issue and propose a novel generative model, $\sigma$-Reg.~CVAE. as the generator $G$.
% 	On three challenging ISCS person Re-ID datasets, CCSFG achieves state-of-the-art results.
    % \textbf{Limitation:}
	\textbf{Potential negative societal impact}: As a more advanced and robust feature learning technique for visual data, the proposed method might be abused for unauthorized monitoring.

\noindent\textbf{Acknowledgement.} \quad
This work was supported by the National Science Foundation for Young Scientists of China (62106289, 62106288), China National Postdoctoral Program for Innovative Talents (BX20200395), China Postdoctoral Science Foundation (2021M693616) and Zhuhai Industry-University-Research Cooperation Project (ZH22017001210010PWC).

	%%%%%%%%% REFERENCES
	{\small
		\bibliographystyle{ieee_fullname}
		\bibliography{egbib}
	}
	
\end{document}